\documentclass[sigconf]{acmart}
\makeatletter
\def\@ACM@checkaffil{
    \if@ACM@instpresent\else
    \ClassWarningNoLine{\@classname}{No institution present for an affiliation}%
    \fi
    \if@ACM@citypresent\else
    \ClassWarningNoLine{\@classname}{No city present for an affiliation}%
    \fi
    \if@ACM@countrypresent\else
        \ClassWarningNoLine{\@classname}{No country present for an affiliation}%
    \fi
}
\makeatother

\usepackage{makecell}
\usepackage{graphicx}
\usepackage{float}
\usepackage{subfig}
\usepackage{multirow}
\usepackage{color}

\AtBeginDocument{%
  }

\newcommand{\reffig}[1]{{Fig.~\ref{fig:#1}}}

\newcommand{\reftab}[1]{{Tab.~\ref{tab:#1}}}
\newcommand{\revised}[1]{}
\definecolor{amber}{rgb}{1.0, 0.49, 0.0}

\newcommand{\todo}[1]{{\color{red}[TODO: #1]}}
\newcommand{\phil}[1]{}

\newcommand{\ignore}[1]{}

\def\ie{\textit{i.e.}}
\def\eg{\textit{e.g.}}

\definecolor{Gray}{gray}{0.85}
\definecolor{White}{gray}{1.0}
\newcolumntype{a}{>{\columncolor{Gray}}c}
\newcolumntype{z}{>{\columncolor{White}}c}

\copyrightyear{2023}
\acmYear{2023}
\setcopyright{acmlicensed}\acmConference[MM '23]{Proceedings of the 31st ACM International Conference on Multimedia}{October 29-November 3, 2023}{Ottawa, ON, Canada}
\acmBooktitle{Proceedings of the 31st ACM International Conference on Multimedia (MM '23), October 29-November 3, 2023, Ottawa, ON, Canada}
\acmPrice{15.00}
\acmDOI{10.1145/3581783.3611839}
\acmISBN{979-8-4007-0108-5/23/10}

\acmSubmissionID{653}

\begin{document}\sloppy

\title[NPF-200: A Multi-Modal Eye Fixation Dataset and Method for Non-Photorealistic Videos]{NPF-200: A Multi-Modal Eye Fixation Dataset and Method for Non-Photorealistic {Videos}}

\author{Ziyu Yang}
\orcid{0009-0006-1077-5944}
\authornote{Both authors contributed equally to this research.}
\affiliation{%
  \institution{South China University of Technology}
}
\email{yangzy.dlut@gmail.com}

\author{Sucheng Ren}
\orcid{0000-0003-4730-8435}
\authornotemark[1]
\author{Zongwei Wu}
\orcid{0009-0008-4568-0560}
\affiliation{%
	\institution{Singapore Management University}
}
\email{
oliverrensu@gmail.com
}
\email{zongweiwu999@gmail.com
}

\author{Nanxuan Zhao}
\orcid{0000-0002-4007-2776}
\affiliation{%
  \institution{Adobe Research}
}
\email{nanxuanzhao@gmail.com
}

\author{Junle Wang}
\orcid{0000-0002-9096-2670}
\affiliation{%
  \institution{Tencent}
}
\email{jljunlewang@tencent.com
}

\author{Jing Qin}
\orcid{0000-0002-7059-0929}
\affiliation{%
  \institution{The Hong Kong Polytechnic University}
}
\email{harry.qin@polyu.edu.hk}

\author{Shengfeng He}
\orcid{0000-0002-3802-4644}
\authornote{Corresponding author.}
\affiliation{%
	\institution{Singapore Management University}
}
\email{shengfenghe@smu.edu.sg}

\renewcommand{\shortauthors}{Ziyu Yang et al.}

\begin{abstract}
Non-photorealistic videos are in demand with the wave of the metaverse, but lack of sufficient research studies. This work aims to take a step forward to understand how humans perceive non-photorealistic videos with eye fixation (\ie, saliency detection), which is critical for enhancing media production, artistic design, and game user experience. To fill in the gap of missing a suitable dataset for this research line, we present NPF-200, the first large-scale multi-modal dataset of purely non-photorealistic videos with eye fixations. Our dataset has three characteristics: 1) it contains soundtracks that are essential according to vision and psychological studies; 2) it includes diverse semantic content and videos are of high-quality; 3) it has rich motions across and within videos. We conduct a series of analyses to gain deeper insights into this task and compare several state-of-the-art methods to explore the gap between natural images and non-photorealistic data. Additionally, as the human attention system tends to extract visual and audio features with different frequencies, we propose a universal frequency-aware multi-modal non-photorealistic saliency detection model called NPSNet, demonstrating the state-of-the-art performance of our task. The results uncover strengths and weaknesses of multi-modal network design and multi-domain training, opening up promising directions for future works. {Our dataset and code can be found at \url{https://github.com/Yangziyu/NPF200}}.
\end{abstract} 

\ccsdesc[500]{Applied computing~Media arts}

\keywords{Non-photorealistic videos, eye fixation, multi-modal frequency}


\begin{teaserfigure}
\centering
  \vspace{-4mm}\subfloat[Data Collection Process]{\includegraphics[width=.41\linewidth]{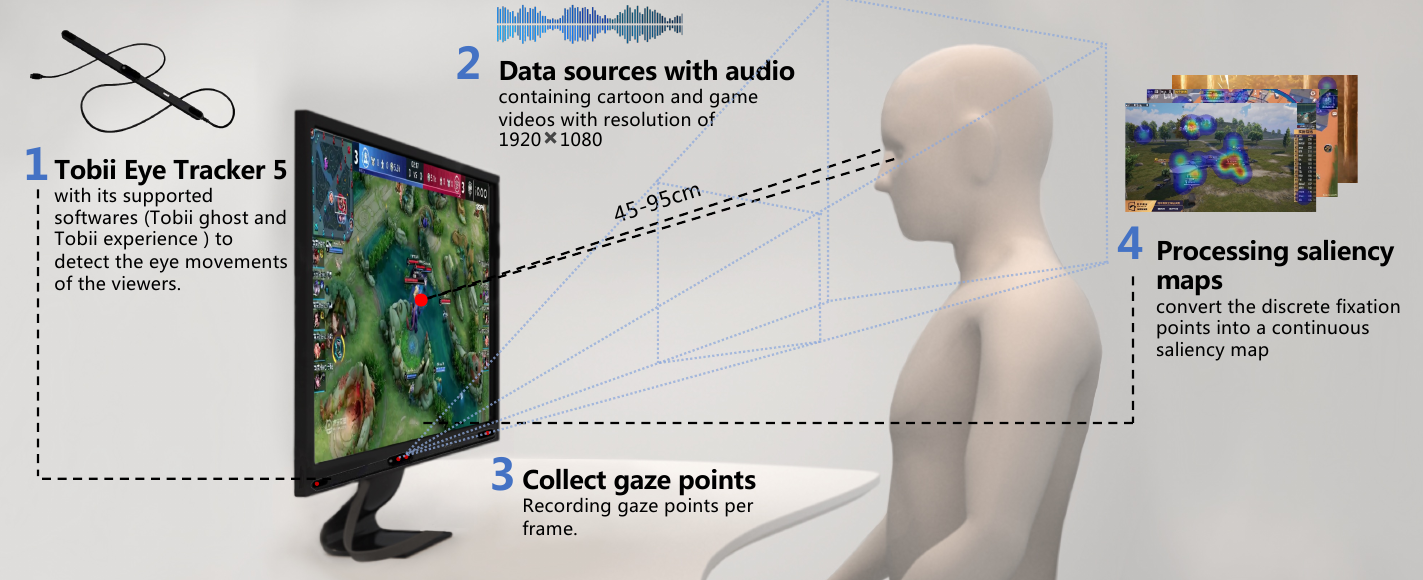}}
  \subfloat[Eye Gaze Movement]{\includegraphics[width=.297\linewidth]{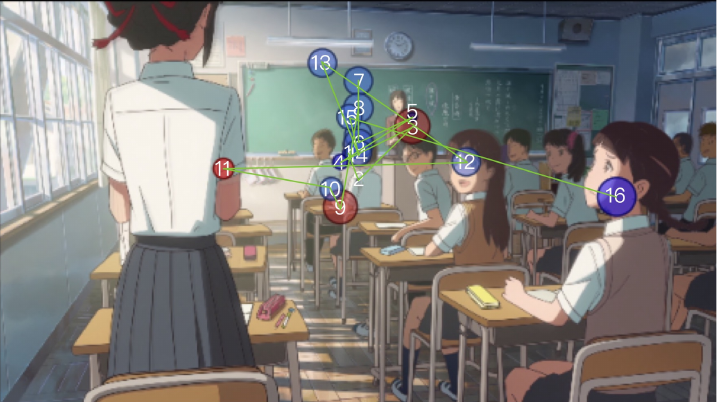}}
  \subfloat[GT (Left) and Our Predictions]{\includegraphics[width=.294\linewidth]{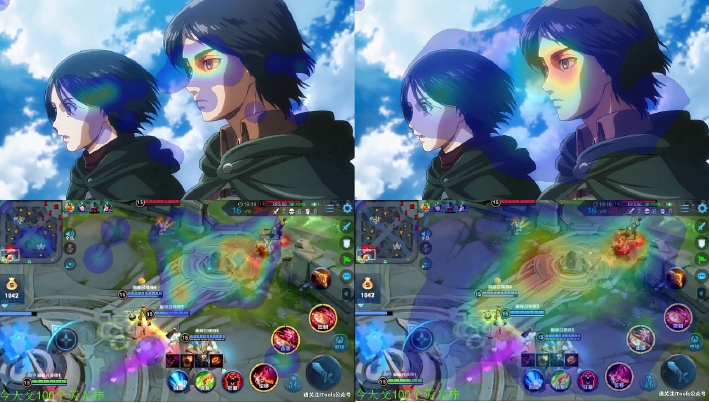}}
  \vspace{-3mm}\caption{We present a multi-modal non-photorealistic eye fixation dataset, NPF-200, obtained by gathering human eye fixation points on non-photorealistic videos that contain a diverse range of content. (a) illustrates the process we used to collect human eye fixation points, and we performed a detailed analysis of eye fixations (b) on non-photorealistic videos. On this basis, we propose a multi-model saliency detection method that encompasses multi-frequency intra-modal and inter-modal feature extractions. The saliency prediction results are depicted in (c).}
  \label{fig:teaser}
\end{teaserfigure}

\maketitle

\section{Introduction}

The wave of the metaverse has spawned a surge of non-photorealistic scenarios, such as virtual game space, animation characters, and graphic design. When perceiving these videos, the human visual system operates in a way by allocating more attention to the interesting regions~\cite{engelke2009visual}. Thus, developing methods for understanding such visual preferences and fixations (\ie,~saliency detection) has practical benefits. By identifying these regions, it becomes possible to quickly extract key scenes or important areas during video editing or media production. Additionally, game and animation designers can use saliency detection to guide viewers' attention and ensure that their intended paths are followed.

Though there are plenty of works targeting saliency detection \cite{wang2018revisiting, wu2020salsac, wang2021spatio}, they mainly study videos with natural scenes, which are not suitable to deal with non-photorealistic videos, as evidenced by psychophysical studies~\cite{sundstedt2008psychophysical, caffrey2008viewer, jie2008video, seif2004visually}. Taking video games as an example, audiences' attention is often directed in a top-to-down manner, following the decision-making during playing. This phenomenon is called Rubicon theory~\cite{gollwitzer1999deliberative}. Additionally, non-photorealistic videos feature unique scene transitions, such as fade-in, fade-out, same-subject transition, and flashback techniques commonly found in cartoons. Besides, non-photorealistic videos not only cover diverse visual appearances with more vivid colors and complex scenes but their dubbing and sound effects also have a great impact on eye movements~\cite{caffrey2008viewer, he2014saliency, doherty2018development}. 

To tackle the above problems, we propose a multi-modal \textbf{N}on-\textbf{P}hotorealistic eye \textbf{F}ixation video dataset called \textbf{\textit{NPF-200}}. Our dataset contains 200 carefully selected videos with more than 70,000 frames, covering diverse categories (\ie, 6 for cartoon and 3 for video games) with various characters, motion patterns, activities, etc. Different from existing saliency datasets on natural videos that discard all audio tracks~\cite{mathe2014actions, mathe2014actions, wang2018revisiting}, inspired by recent psychological studies \cite{perrott1990auditory, vroomen2000sound}
	and eye-tracking analyses \cite{song2013different, min2014sound, he2016exemplar, chen2022comprehensive, he2019exploring}, we preserve audios which are essential auxiliary information for saliency detection. To the best of our knowledge, NPF-200 is the largest eye fixation dataset for multi-modal non-photorealistic videos. 

On the basis of our NPF-200 dataset, we conduct a series of analyses on our dataset for gaining more insights and understanding, and build a benchmark by examining current state-of-the-art saliency detection methods. In the context of analyzing the NPF-200 dataset for saliency detection methods, understanding how the human visual and auditory systems process information at different frequencies can inform the development of more effective algorithms. As the human visual and auditory system has been observed to extract visual elementary features at varying frequencies~\cite{bullier2001integrated, he2017delving, saaty2022game}. Low frequencies tend to provide global information about a given visual stimulus, such as objects of considerable size that consistently appear throughout a video, as well as prolonged sounds like the footsteps of the character. Conversely, high frequencies convey local spatial changes in the visual stimulus, typically related to objects of smaller sizes and with vast temporal variance in videos, and also informative audio clips with short durations like the brief gunshot. For example, it is common in first-person shooter games for there to exist a notable size discrepancy between the target characters and the protagonist, yet the human visual system is able to capture information from both simultaneously. These observations suggest that the human visual system processes visual information in a frequency-dependent manner, with different frequencies carrying distinct information about the visual stimulus.

In light of the frequency-dependent processing of visual and auditory information in the human attention system, we propose \textbf{NPSNet} (Non-Photorealistic Saliency Network), that encompasses multi-frequency intra-modal and inter-modal feature extraction. Central to the effectiveness of this method is the Universal Frequency-aware Module, which enables the extraction of intricate and comprehensive relationships between video and audio features, thereby paving the way for a more sophisticated and nuanced understanding of the underlying mechanisms governing saliency detection in non-photorealistic videos.

The outcomes of our experiments offer an in-depth understanding of the results and point towards promising directions for multi-modality and multi-domain training, further contributing to the development and comprehension of non-photorealistic design. Our benchmark includes crucial codes and tools that enable researchers to conduct a thorough evaluation of their methods and thus propel the field of non-photorealistic saliency detection forward. We will share our dataset and model upon acceptance to ensure accessibility and promote further research in the field.

 \section{Related Work}
There exist various video datasets for saliency detection, as outlined in \reftab{related}. They can be categorized into: 1) visual-only, by extracting key information from the visual scenes; 2) visual-audio, with multi-modal resources; 3) eye-tracking datasets that collect human fixation points specifically for non-photorealistic videos.

\begin{table}[t]
    \setlength\tabcolsep{0.6pt}
    \caption{Statistics of eye-tracking datasets. ``Free-view'' indicates the viewing method adopted by the participants, while ``multiple fixations'' indicates whether each frame contains fixation points from multiple volunteers.}\vspace{-3mm}
		\resizebox{\columnwidth}{!}{%
    \begin{tabular}{l|c|c|c|c|c|cc|c|c}
    \hline
    \multicolumn{1}{l|}{\multirow{2}{*}{\textbf{Dataset}}} & \multirow{2}{*}{Year} & \multirow{2}{*}{Videos} & \multirow{2}{*}{Viewers} & \multirow{2}{*}{\begin{tabular}[c]{@{}c@{}}Free \\View \end{tabular}} & \multirow{2}{*}{Audio} & \multicolumn{2}{c|}{NPF}    & \multirow{2}{*}{\begin{tabular}[c]{@{}c@{}}Multiple \\Fixations \end{tabular}} & \multirow{2}{*}{dynamic} \\ \cline{7-8}
    \multicolumn{1}{c|}{}&  & &  &  & & \multicolumn{1}{c|}{cartoon}  & games  &  &   \\ \hline
    Hollywood-2 \cite{mathe2014actions}& 2012 & 1707  & 19  &   &  & \multicolumn{1}{c|}{}  &  & $\checkmark$ & $\checkmark$    \\
    UCF sports \cite{mathe2014actions}   & 2012 & 150   & 19   &   &   & \multicolumn{1}{c|}{}  &   & $\checkmark$ & $\checkmark$ \\
    DIEM \cite{mital2011clustering}  & 2011 & 84    & $ \sim $ 50  & $\checkmark$ & $\checkmark$  & \multicolumn{1}{c|}{}  &  & $\checkmark$  & $\checkmark$   \\
    DHF1K  \cite{wang2018revisiting}  & 2017 & 1000  & 17    & $\checkmark$  &   & \multicolumn{1}{c|}{}  &  & $\checkmark$ & $\checkmark$  \\
    Coutrot1  \cite{coutrot2014saliency} & 2014 & 60   & 18    & $\checkmark$    & $\checkmark$  & \multicolumn{1}{c|}{} & & $\checkmark$ & $\checkmark$ \\
    SumMe \cite{gygli2014creating},\cite{tsiami2019behaviorally}  & 2014  & 25       & 15  & $\checkmark$  & $\checkmark$  & \multicolumn{1}{c|}{}  &  & $\checkmark$  & $\checkmark$ \\
    Coutrot2 \cite{coutrot2015efficient}  & 2015 & 20   & 15   & $\checkmark$   & $\checkmark$  & \multicolumn{1}{c|}{}  & & $\checkmark$ & $\checkmark$  \\
    ETMD \cite{koutras2015perceptually}  & 2015 & 12 & 10    & $\checkmark$    & $\checkmark$   & \multicolumn{1}{c|}{}   &   & $\checkmark$  & $\checkmark$   \\
    AVAD \cite{min2016fixation} & 2016 & 45  & 14  & $\checkmark$  & $\checkmark$  & \multicolumn{1}{c|}{}   &   & $\checkmark$   & $\checkmark$   \\
    Atari-HEAD \cite{zhang2020atari} & 2019  & $\sim$468  & 4    &   &    & \multicolumn{1}{c|}{}   & $\checkmark$ &    & $\checkmark$    \\
    Cat2000 \cite{borji2015cat2000} & 2015  & images &  &    &    & \multicolumn{1}{c|}{$\checkmark$} &    &   &     \\
    \textbf{NPF-200 (Ours)}  & 2022  & 200   & 20   & $\checkmark$  & $\checkmark$  & \multicolumn{1}{c|}{$\checkmark$} & $\checkmark$ & $\checkmark$ & $\checkmark$  \\ 
	\hline
    \end{tabular}%
    }
    \label{tab:related}
    \vspace{-2mm}\end{table}

\textbf{Visual datasets.}
The Hollywood-2 dataset \cite{mathe2014actions} is a collection of 1,707 videos from the Hollywood-2 action recognition dataset \cite{hadizadeh2011eye}. This dataset includes fixation points from 19 observers, but only 3 of them were free-viewing, while the rest were mission-driven viewers. It is primarily focused on human actions and movie scenes. The UCF sports fixation dataset \cite{mathe2014actions} contains 150 videos from the UCF sports action dataset \cite{rodriguez2008action}, which includes realistic sports videos from broadcast television channels. This dataset covers 9 common sports action classes but, similar to the Hollywood-2 dataset, is also biased towards task-aware observation. The DHF1K dataset \cite{wang2018revisiting} is a large-scale gaze dataset collected during free-viewing of 1,000 YouTube videos classified into seven main categories. The fixation points from 17 observers were collected to generate saliency maps. However, all these datasets contain only a single visual modality, and their scope is limited to specific domains or tasks.

\textbf{Visual-audio datasets.}
The DIEM dataset \cite{mital2011clustering} collected data from more than 250 participants watching 85 different video shots, sourced from publicly accessible repositories and a handful of game trailers. Each video contains free-viewing fixations from 42 observers. The Coutrot database \cite{coutrot2014saliency, coutrot2015efficient} is divided into Coutrot1 and Coutrot2. Coutrot1 contains 60 clips with dynamic natural scenes split into four visual categories: one/several moving objects, landscapes, and faces. Coutrot2 contains 15 videos of four people having a meeting. The corresponding eye-tracking data is collected from 40 persons for conducting free-viewing. The AVE dataset \cite{tavakoli2019dave} consists of 150 hand-picked video sequences from the DIEM, Coutrot1, and Coutrot2 datasets. The videos are classified into three categories - Nature, Social Events, and Miscellaneous. Although these datasets contain both visual and audio modalities, they mainly serve for studying natural scenes. In contrast, we target non-photorealistic videos and collect a large-scale dataset with multi-modal labels for comprehensive research on this novel task.

\textbf{Non-photorealistic datasets.}
The Atari-HEAD dataset \cite{zhang2020atari} and other similar datasets, such as those proposed by \cite{alkan2007studying, kenny2005preliminary, sundstedt2008psychophysical}, provide eye-tracking data for different gaming scenarios. However, these datasets have a specific collection goal and only include eye-tracking data from a single player during gameplay. This limitation makes it difficult to generalize the data and apply it for saliency prediction. In contrast, the NPF-200 dataset proposed in this work contains eye-tracking data from 20 volunteers for each frame and includes three common types of games on the market. This diversity provides a more comprehensive range of stimuli for saliency prediction tasks. Additionally, most cartoon eye-tracking datasets only include data from images, such as Cat2000 \cite{borji2015cat2000}, which lacks temporal information during cartoon playback. However, saliency prediction from images cannot be easily generalized to videos, as shown in \cite{nguyen2013static, wang2019revisiting}. The proposed NPF-200 dataset bridges this gap by providing eye-tracking data for non-photorealistic videos, which can aid in the development of effective saliency detection.

\textbf{Visual-only Saliency Detection.}
Early methods predict video saliency \cite{harel2006graph, sultani2014human} based on low-level visual statistics
and additionally consider temporal features compared to conventional static models \cite{le2006coherent, zhang2008sun, bruce2005saliency}.
The performance of these early models is limited by the ability of the low-level features to represent temporal information.
Recently, deep learning based methods \cite{zhang2018video, jiang2018deepvs, ren2020tenet} have been introduced for video saliency detection,
exploring different strategies to extract temporal features (\eg, optical flow, LSTM, 3D convolutions, convGRU) \cite{wang2020learning, niu2020boundary}.
Bak et al.\cite{bak2017spatio} propose a two-stream network to process video frames and optical flow maps simultaneously.
Wang et al.\cite{wang2018revisiting} and Wu et al. \cite{wu2020salsac} propose an attention mechanism with ConvLSTM to achieve better performance.
Droste et al. \cite{droste2020unified} and Lai et al. \cite{lai2019video} model long-term temporal characteristics by convGRU.
TASED-Net \cite{min2019tased} adopts S3D \cite{xie2018rethinking} as encoder and uses 3D convolutions as decoder, achieving appealing
performance. HD2S \cite{bellitto2021hierarchical}, ViNet \cite{jain2021vinet} ,TSFP \cite{chang2021temporal} and STSANet \cite{wang2021spatio} also build model based on 3D convolutions.
Among these, \cite{wang2021spatio,wu2020salsac,ren2021reciprocal} take advantage of the self-attention mechanism to capture long-range relations between spatiotemporal features.
VSFT \cite{ma2022video} further proposes a transformer-based architecture for visual saliency detection.

\textbf{Visual-audio Saliency Detection.}
The influence of audio on human visual attention has been widely investigated and proven by many psychological studies \cite{perrott1990auditory, vroomen2000sound}
and eye-tracking analysis \cite{song2013different, min2014sound}. Recent studies have begun to explore fusing visual and audio modality information for saliency detection \cite{wang2021semantic, min2020multimodal}.
SoundNet \cite{aytar2016soundnet} leverages the natural synchronization between vision and sound to learn an acoustic representation.
STAViS \cite{tsiami2020stavis} extends the SUSiNet \cite{koutras2019susinet} visual saliency model, combining spatio-temporal visual and auditory information to
efficiently predict visual-audio saliency. AViNet\cite{jain2021vinet} and TSFP \cite{chang2021temporal} combine the information captured from visual and audio signals to further enhance the saliency detection performance.
Limited by the number and variety of visual-audio saliency datasets, visual-audio saliency detection models are not well developed. To remedy this shortcoming, we propose
a new dataset containing video clips and corresponding audio, called NPF-200. Based on this dataset, we propose NPSNet to achieve state-of-the-art visual-audio saliency detection.

\section{NPF-200 Dataset}\label{dataset}
We introduce NPF-200, an eye-tracking-based multimodal saliency detection dataset for non-photorealistic videos. We select mainly two categories that are commonly appeared in non-photorealistic videos: cartoon videos for 2D scenes and game videos for 3D scenes.
As shown in \reftab{NPF200}, Our dataset contains 200 videos with audio of various virtual scenes, content, and duration, including 100 cartoon videos and 100 game videos.
Each video is accompanied by eye-tracking saliency maps collected from 20 observers.
   
\textbf{Device for Collection.}
We use Tobii Eye Tracker 5 with its supported softwares (\ie, Tobii ghost and Tobii experience ) to detect the eye movements of the viewers.
Specifically, the video is displayed on a 27-inch display screen with a display resolution of 1920$\times$1080, while the console records the viewer’s gaze at a sampling rate of 60Hz.
According to the product manual book, we require the viewer's head to be kept at a horizontal distance of 45-95 cm from the eye tracker.

\textbf{Data Sources and Participants.}
Our videos are downloaded from Youtube and Bilibili. The resolution of cartoon videos is 1080p, and the resolution of game videos is 720p.
We invited 20 people who major in computer science and have no eye-tracking experience before to participate in the collection of eye-tracking saliency data. Ten of them were men and ten were women, and they all age between 20 and 29.
We have informed participants of the purpose of collecting the saliency data and obtained the consents from them.

\begin{figure}[t]
	\centering
        \vspace{-3mm}
	\captionsetup[subfloat]{justification=centering,labelformat=empty}
	\subfloat[]{
		\begin{minipage}{\linewidth}				
			\includegraphics[width=.9\linewidth]{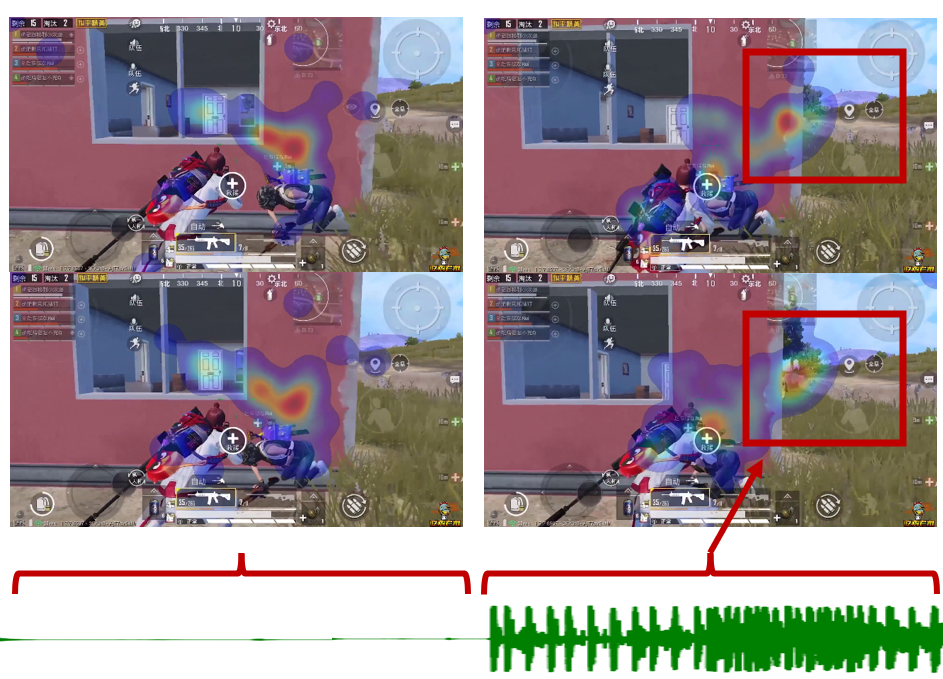}\\
			\centering{\small{\text(a) Audio can guide human attention to focus on key areas.}}\\
	        \hspace{-0.5cm}
            \includegraphics[width=.9\linewidth]{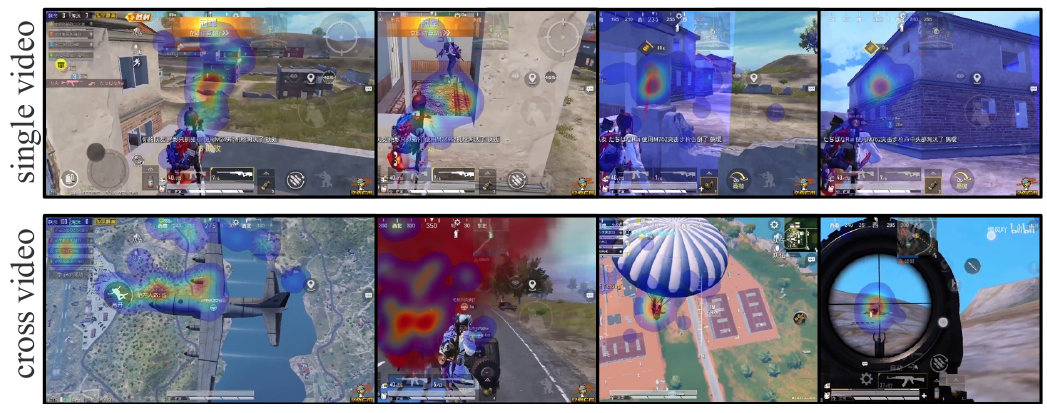}\\
			\centering{\small{\text(b) Diverse semantic contents.}} \\
            \includegraphics[width=.45\linewidth]{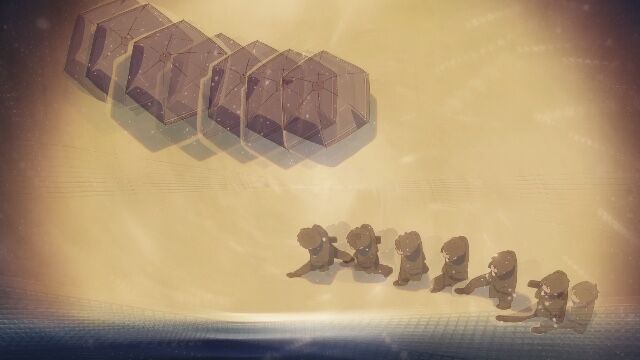}
            \includegraphics[width=.45\linewidth]{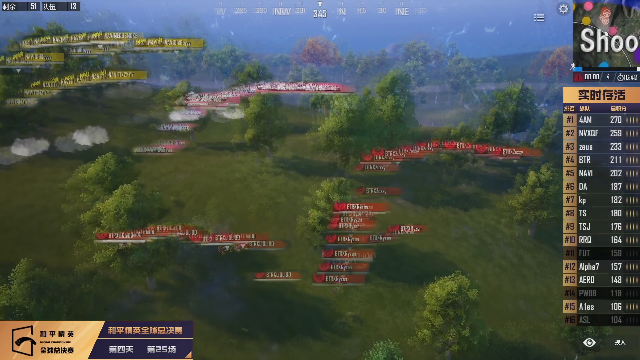}\\
			\centering{\small{\text(c) Rich motions.}}\\

	\end{minipage}
	}	
	\vspace{-7mm}\caption{Characteristics of our NPF-200 dataset.}\vspace{-5mm}
\label{fig:datasetvisualize}
\end{figure}

\subsection{Data Collection Process}

Before collecting eye-tracking data, we asked participants to align the tracking device by looking at six fixation points in the middle and edges of the screen. For near-sight participants, we asked them to wear glasses for vision correction. Participants did not know the type and content of the videos before watching them.
Videos are shown to each viewer in a random order with 5-second breaks between videos to prevent the viewer from losing concentration due to eye strain.
The eye-tracking device records the positional value of the observers' attention in each frame. Finally, we collected fixation points from 20 observers in each frame from 200 videos.
To convert the discrete fixation points into a continuous saliency map, we blur each fixation points with a small Gaussian kernel, following the same approach in \cite{wang2018revisiting}.
The final saved continuous saliency map is normalized to a range between 0 and 1.
Note that the eye-tracking device initially collects fixation maps. The saliency map is the result of Gaussian blurring based on the coordinates of the points on the fixation map.
We use the saliency map for both training and testing.

\textbf{Dataset split.}
We split NPF-200 into a training set and a test set, where the training set contains 82 game videos and 82 cartoon videos, and the test set contains 18 game videos and 18 cartoon videos.
Note that the number of different types of cartoons and games in the test set is uniformly and randomly selected, to avoid containing only a certain kind of game or cartoon.

\subsection{Dataset Characteristics}

As the \textbf{first} large-scale multi-modal eye-fixation (saliency) dataset for non-photorealistic videos, our dataset mainly contains the following characteristics:

\textbf{Audio-aided.}
Scientific research has proved that audio affects people's visual attention \cite{saaty2022game}, but most of the existing video saliency datasets only have videos without soundtracks.
Especially in virtual scenes, the importance of audio is more obvious than in real scenes. Audio can provide audiences with more effective content guidance in cartoons and games.
For example, as shown in \reffig{datasetvisualize}, the audio of gunshots and footsteps in shooting games has guided the viewer's visual attention.
Therefore, all the videos in our NPF-200 are accompanied by audio data to assist in saliency detection.

\begin{table}[t]
    \setlength\tabcolsep{1.2pt}
    \caption{
        Number of frames in each category of the NPF-200 dataset, where ``WWY'' stands for ``Weathering With You'', ``AoT'' for ``Attack on Titan'', ``DS'' for ``Demon Slayer``, ``YN'' for ``Your Name'', ``HMC'' for ``Howl's Moving Castle'', ``SA'' for ``Spirited Away'', ``GI'' for ``Genshin Impact'', ``HoK'' for ``Honor of Kings'', and ``GfP'' for ``Game for Peace''.
    }\vspace{-3mm}
    \resizebox{\columnwidth}{!}{%
    \begin{tabular}{@{}l|cccccc|cccc@{}}
    \toprule
    \multirow{2}{*}{NPF-200} & \multicolumn{6}{c|}{Cartoon}                                                                                                                     & \multicolumn{3}{c}{Games}                                       \\ \cline{2-10} 
                         & \multicolumn{1}{c|}{WWY}  & \multicolumn{1}{c|}{AoT}  & \multicolumn{1}{c|}{DS}   & \multicolumn{1}{c|}{YN}   & \multicolumn{1}{c|}{HMC}  & SA   & \multicolumn{1}{c|}{GI}    & \multicolumn{1}{c|}{HoK}   & GfP   \\ \hline
Frames                   & \multicolumn{1}{c|}{7551} & \multicolumn{1}{c|}{5716} & \multicolumn{1}{c|}{6241} & \multicolumn{1}{c|}{6996} & \multicolumn{1}{c|}{5243} & 7079 & \multicolumn{1}{c|}{12030} & \multicolumn{1}{c|}{13933} & 10043 \\ \hline
Duration                & \multicolumn{6}{c|}{8-25(s)}                                                                                                                     & \multicolumn{3}{c}{10-30(s)}                                    \\ \hline
Training Videos          & \multicolumn{6}{c|}{82}                                                                                                                          & \multicolumn{3}{c}{82}                                          \\ \hline
Testing Videos           & \multicolumn{6}{c|}{18}                                                                                                                          & \multicolumn{3}{c}{18}                                          \\ \hline
\end{tabular}%
    }
    \label{tab:NPF200}
    \vspace{-3mm}\end{table}
    
\textbf{Diverse semantic content and high quality.}
Our videos are widely derived from different cartoon and game videos, and cover diverse scenes and semantic contents. Because of the creative nature of non-photorealistic videos, the mentioned characters, interactions, background, and many other factors vary a lot, which also have really different appearance distributions from natural videos. We show a few examples in  \reffig{datasetvisualize}. As can be seen, the environments and landscapes in shooter games are varied and contain a lot of virtual guidance such as anchors, maps, and control pads. Moreover, the videos in our dataset are of high quality. We remove all the watermarks and subtitles in videos to avoid them affecting the viewer's attention.
All videos are stored in \textit{.mp4} format at 30fps and uniformly resized to a resolution of 640×360. The audio is uniformly stored in two-channel \textit{.wav} format.
In summary, our dataset consists of 74,832 video frames (the distribution is shown in \reftab{NPF200}  with a total duration of 2,514 seconds.

\textbf{Rich motions.}
The idea that motions can influence human attention has been confirmed in existing computer vision and cognition research \cite{mital2011clustering, jiang2017exploiting}. To take this into consideration, our dataset also embraces diverse motions, such as camera motion and content motion. Some of the examples have shown in \reffig{datasetvisualize}.

We believe that the above characteristics in total not only provide rich information for dataset analysis to facilitate cognitive studies but also enrich the modal information of virtual scenes for better durable performance and benchmark.

\begin{figure}[t]
	\centering
			\includegraphics[width=.49\linewidth]{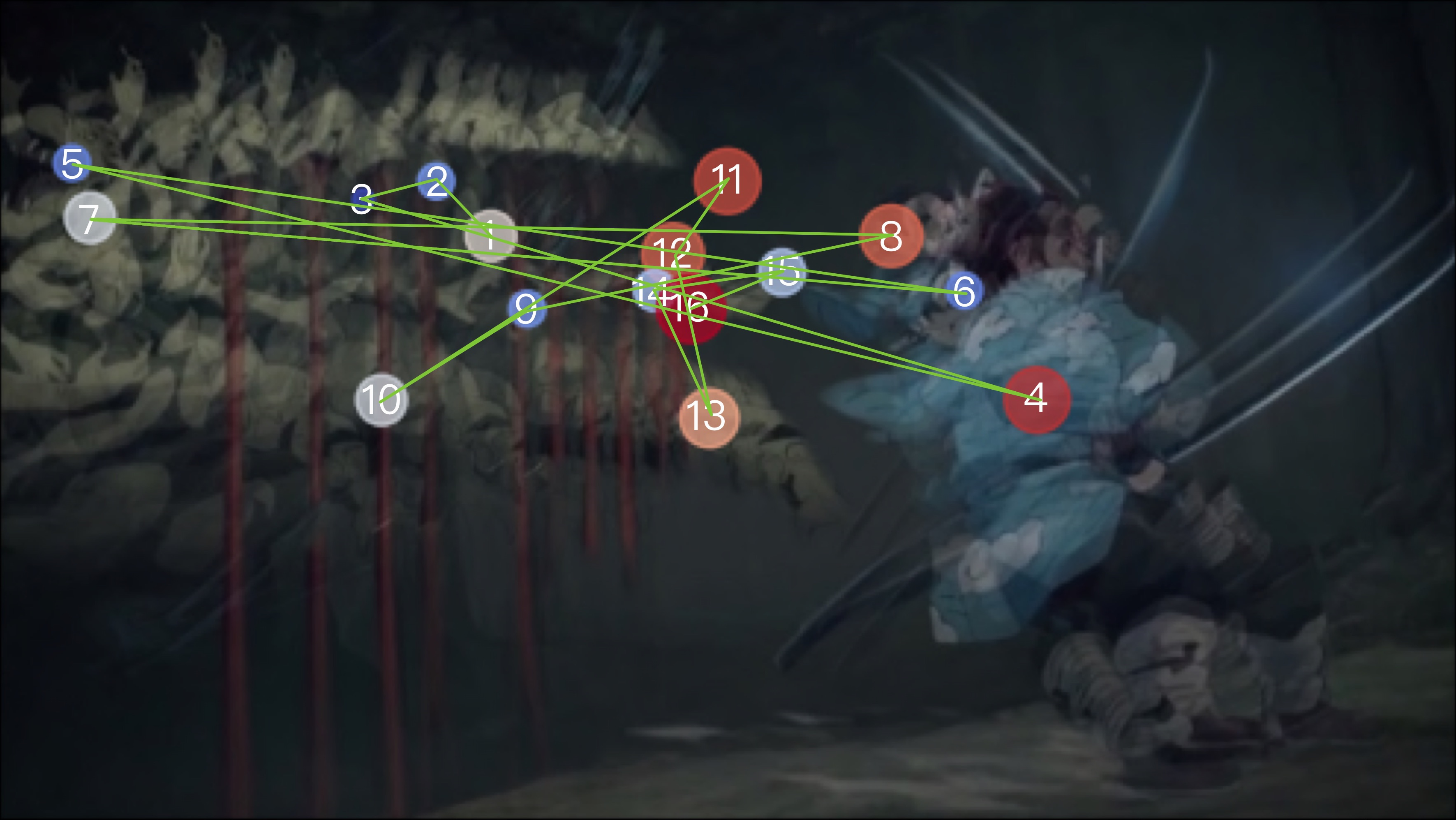}
			\includegraphics[width=.49\linewidth]{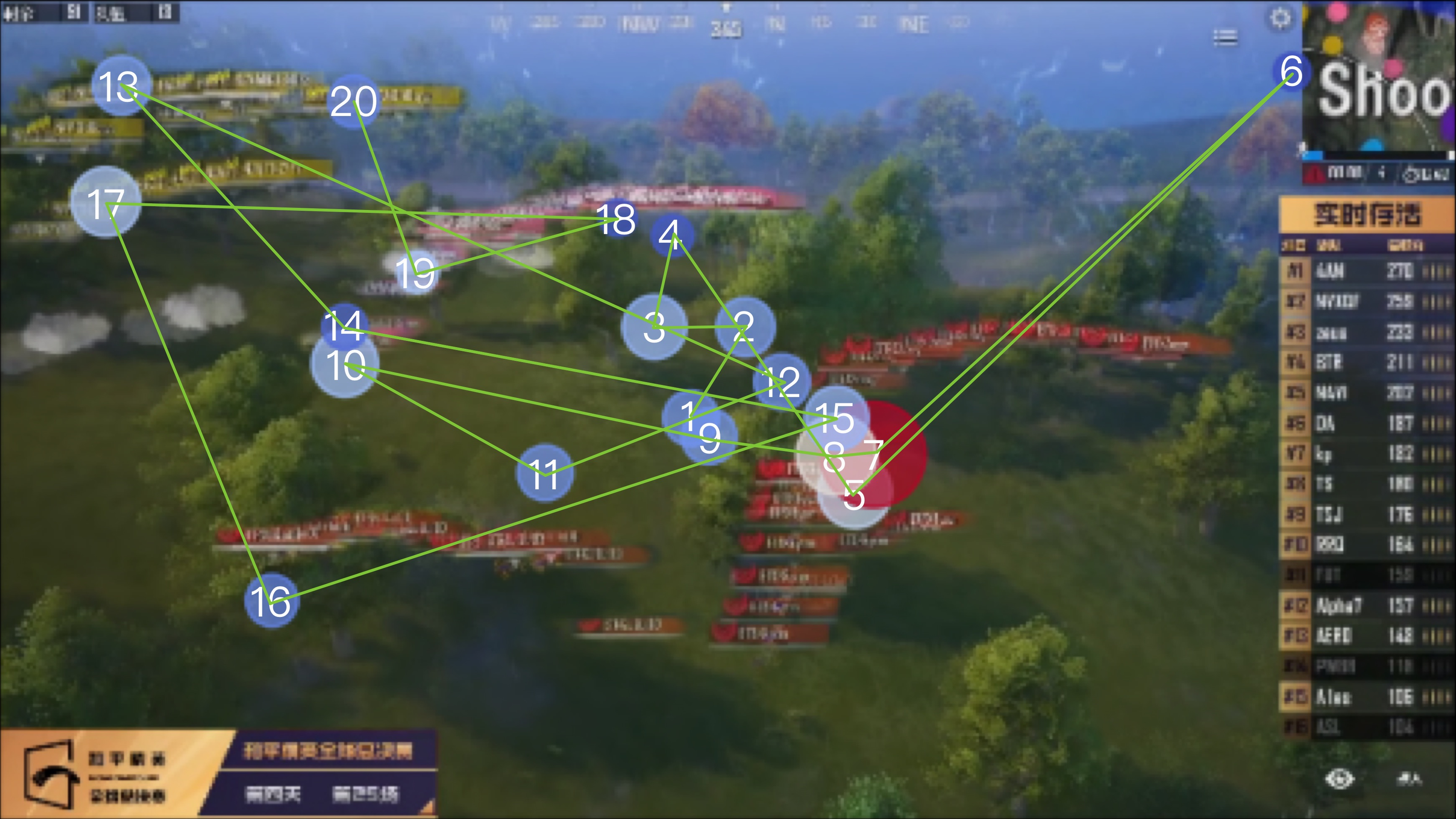}
	\vspace{-2mm}\caption{Eye gaze movement points of example video clips from our dataset in successive video frames. The color and size of each point illustrate its saliency proportion in the map with warmer and larger indicating more importance.}\vspace{-4mm}
	\label{fig:gaze}
\end{figure}

\subsection{Analysis on Video Eye Gaze Movement}

To understand how humans watch non-photorealistic videos, we visualize human eye movement
in successive video frames of our dataset into a single image as shown in \reffig{gaze}.
This image contains the motion of multiple subjects and scenes.
Small dots are used to mark the human eye gaze points in this image,
and the number on the dot indicates the frame number corresponding to the current eye gaze points.
The size and color of the dot represent the saliency proportion of this fixation point in
its saliency map. The larger and the warmer color (\ie, redder) of the dot,
the higher the saliency proportion of this fixation point.
Conversely, the smaller and colder color (\ie, bluer) of the dot, the lower the significant proportion.
We connect the points sequentially in the order of frames to explore human eye movement.
From the cartoon eye movement image on the left, we can see that humans tend to focus on moving objects.
When there are multiple moving objects and the distance is not far,
the human eye gaze will switch back and forth between multiple objects, then
most of the gazes will be within a range where they can see multiple moving objects
at the same time. This phenomenon is also reflected in the game eye movement image.
In addition, in non-photorealistic game videos, as shown on the right,
humans pay attention to instructive information such as game maps, player names, and smoke. We believe there are more interesting findings to be uncovered in our dataset, and we encourage further exploration of these patterns in future research.

\section{Frequency-aware Fixation}\label{method}

\begin{figure*}[t]
	\centering
	\includegraphics[width=0.95\linewidth]{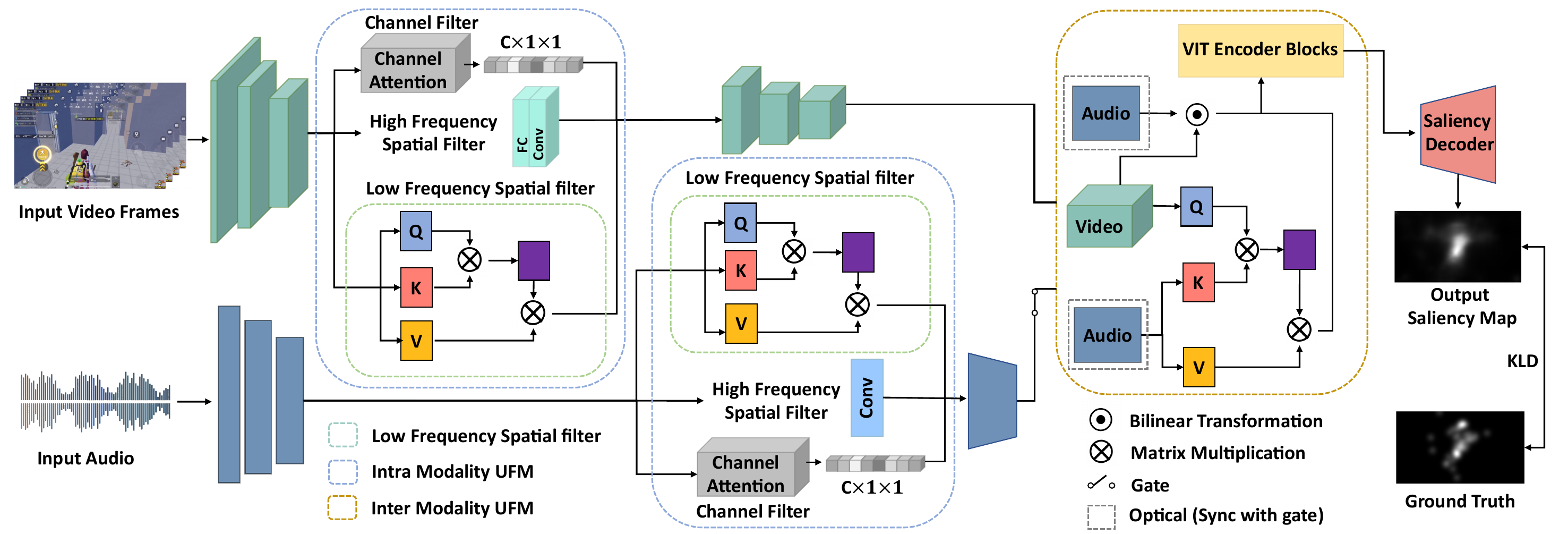}
	\vspace{-3mm}\caption{Pipeline of our NPSNet with the input of the current frame, its visual context (preceding $k-1$ frames), and corresponding audio waveform of these $k$ frames to predict the saliency map. UFM denotes the Universal Frequency-aware Module.}\vspace{-2mm}
	\label{fig:method}
\end{figure*}

Our approach aims to predict the saliency map $M_i$ of the $i$-th frame by incorporating multi-modality context to enhance the model's capacity for capturing temporal and auxiliary information. The multi-modality context includes two types of information: 1) visual context represented by the preceding $k-1$ frames and 2) audio context denoted by the corresponding soundtrack of the encompassed $k$ frames. Fig. \ref{fig:method} shows the pipeline of our model.

To extract intra-modality features, we utilize the VideoNet and AudioNet, which respectively take in video and audio inputs. To capture multi-frequency features from these modalities, we equip these networks with a Universal Frequency-aware Module (UFM). After extracting features from both modalities, we perform an interaction process that identifies and emphasizes inter-modality features with multi-frequency correspondence. These features are then processed with an Inter-modality Universal Frequency-aware Module, which further enhances their discriminative power.

\subsection{Intra-modality Feature Extraction}
To extract multi-frequency intra-modality features, we propose the incorporation of a \textbf{Universal Frequency-aware Module}, designed to capture essential information across various frequencies. This module is seamlessly integrated into both the \textbf{VideoNet}, which processes video frames as input, and the \textbf{AudioNet}, which handles audio waveforms as input. Through this innovative integration, the Universal Frequency-aware Module effectively enhances the capability of each network in discerning intricate patterns and relationships within their respective modalities, ultimately leading to a more robust and comprehensive representation of the data.

\textbf{Universal Frequency-aware Module.}  To accurately and comprehensively model these features in both audio and video formats, we propose the Universal Frequency-aware Module, which is capable of capturing both high- and low-frequency information. The proposed module consists of specially designed high-frequency filters that are capable of capturing information across short ranges, such as small salient objects or objects that undergo rapid and drastic variations in a short time span \cite{park2022vision}:
\begin{equation}
    \begin{split}
        \hat{X}_i &= FC({X}_i; \theta), \\
        X_i^h &= Conv(\hat{X}_i; \theta)
    \end{split}
\end{equation}
where $FC(\star; \theta)$ is fully connected layer and $Conv(\star; \theta)$ is depth-wise convolution. {$X_i$ represents the input feature of the UFM, which is the input feature of the i-th layer of the encoder.}

The low-frequency filters are in parallel with high-frequency filters to capture the information across long ranges in the feature map like large salient objects and segments of audio with long lengths that contain significant information \cite{park2022vision}:
\begin{equation}
    X_i^l = MSA(X_i)
\end{equation}
where $MSA(\star)$ is the multi-head self-attention~\cite{vaswani2017attention}.

The above filters are for spatial information, we design the third branch for channel information filtering :
\begin{equation}
    \begin{split}
        e_i &= FC(Pool(X_i);\theta), \\
        X_i^c &= e_i\odot X_i
    \end{split}
\end{equation}
where $Pool(\star)$ is the pooling operation. By incorporating such advanced techniques into our models, we aim to establish a universally effective approach to the analysis of eye fixation in both audio and video formats. This would contribute significantly to the development of more sophisticated and robust models for analyzing visual attention and perceptual processing. This module is placed in the video and audio net to enhance the extraction of multi-frequency features.


\textbf{AudioNet.}
We adopt seven layers SoundNet~\cite{aytar2016soundnet} as the architecture of our AudioNet with a universal frequency-aware module at the fifth layer. Benefiting from transfer learning, this model is pretrained on audio-visual scene classification task to learn to better extract audio representation. The audio waveform first be preprocessed by a Hanning window for providing higher weight to the interested time instance before feeding into the AudioNet to obtain the audio features.

\textbf{VideoNet.}
To efficiently extract the temporal and spatial visual features of video, we select the video swin transformer\cite{liu2022video} model as the architecture of our VideoNet with a universal frequency-aware module at the third stage. It accepts three-dimensional video frame input including the time dimension and uses swin transformer block and patch merging to simulate the convolutional neural network to continuously reduce token proficiency and increase the receptive field of each token to reduce the amount of calculation and improve performance. In order to provide VideoNet with temporal and spatial prior knowledge of video, we load the video swin transformer parameters pretrained on Kinetics dataset\cite{kay2017kinetics} for action recognition into VideoNet.
We extract the visual feature maps by sending the video frames to VideoNet.
Meanwhile, we also extract feature maps from the shallow and deep layers of VideoNet, which contain low-level structural features and high-level semantic features, respectively.
These features are used for long-range skip concatenation to aid saliency map detection.

\subsection{Visual-Audio Interaction} 
With intra-modality video and audio features, we explored how to obtain temporal and spatial visual features and corresponding audio features. However, there is an essential gap between video and audio modality.
The goal of this module is to fuse video and audio modality with inter-modality correspondence.

\textbf{Inter-modality correspondence.}
Audio features and visual features extracted from a video clip belong to two modalities, but they have implicitly associated semantic information. Specifically, the audio must be aligned with the video frames in order to achieve audio-video synchronization.
For example, in games and cartoons, the video frames of the characters moving should correspond to the audio of footsteps; in the shooting game, the shooting should have corresponding visual and sound effects during combat.
Therefore, we need to explore how to bridge the gap and construct the correspondence between these two modalities, that is, to locate the position of the audio in the video frames.
To connect these two modalities, we utilize an inter-modality universal frequency-aware module with the high-frequency filter and low-frequency bilinear transformation to construct the inter-modality correspondence of the audio features and visual features. 
Concretely, given the audio feature vector $h_{a} \in \mathbb{R}^{k \times d_{a}}$ and video feature vector $h_{v} \in \mathbb{R}^{k \times d_{v}}$, we calculate inter-modality correspondence through bilinear transformation as follows:

\begin{equation}
    \begin{aligned}
        A_l &=h_{a}^{T}W h_{v}+b,~~
        A_h = MCA(h_{v}, h_{a}, h_{a}),\\
        A &= A_l+A_h
    \end{aligned}
  \end{equation}
  where $W \in \mathbb{R}^{d_{a} \times d_{o} \times d_{v}}$ and $b \in \mathbb{R}^{d_{o} \times 1} $ are learning parameters, $d_a, d_v, d_o$ are the dimensions of $h_a, h_v, A$, respectively. {The $MCA(\star)$ is the multi-head cross-modal attention.} And we obtain the video-audio affinity vector $A \in \mathbb{R}^{k \times d_{o}}$, indicating inter-modality relation. 

To enhance the inter-modality correspondence, we select the VIT\cite{dosovitskiy2020image} encoder, which is good at resolving long-range dependencies and temporal relations, as the architecture of our self-attention encoder.
And we load VIT parameters pretrained on the kinetics dataset\cite{kay2017kinetics} to provide video temporal and spatial priors.
Specifically, since the transformer architecture is permutation-invariant\cite{carion2020end}, we add the video-audio affinity vector and a fixed positional encoding to the self-attention encoder.
The output of our self-attention encoder goes through a 3D convolution and a 3D adaptive average pooling to obtain the saliency feature map. 

\subsection{Saliency Map Prediction}
The saliency decoder adopts six 3D convolutional layers combined with trilinear upsampling to map the saliency features to the output saliency map. Besides, the object position of the input video frames and the output saliency map are spatially correlated, and the structural features and semantic features we extracted from the shallow and deep layers of VideoNet contain the information of the object position.
So we concatenate these  feature maps into the output of each layer in the middle of the saliency decoder and then input it to the next layer. Finally we obtain the saliency map, which indicates the area of people's visual attention of the input interested frame.

\subsection{Loss Function}\label{sec:train_test}
%

We train our model with the supervision signal provided by the saliency maps from the NPF-200 dataset. 
To guide NPSNet to predict reasonable saliency maps, we use KLD (kullback-leibler divergence) 
as the training loss function to measure the difference between the two probability distributions 
$P$ (predicted saliency map $P \in [0,1]^{h,w}$) and $Q$ (ground truth saliency map $Q \in [0,1]^{h,w}$). 
The KLD loss function can be defined as follows:
\begin{equation}
  \begin{aligned}
    \mathcal{L}_{KLD}(W|P,Q)=\sum_{x,y}Q(x,y;W)log \left(\epsilon + \frac{Q(x,y;W)}{P(x,y;W)+\epsilon}\right) 
\text{,}
  \end{aligned}
\end{equation}
where $W$ is the training parameters, and $\epsilon$ is for regularization. 

\section{Experiments}\label{experiment}

\begin{table*}[]    
    \caption{Comparison of various models on different scenes. The models marked without special symbols are trained on DHF1K, and models marked with * are separately trained on game and cartoon, while models containing audio are marked with @.}\vspace{-3mm}
    \resizebox{\textwidth}{!}{%
    \begin{tabular}{c|ccccc|ccccc|cllll}
    \hline
    \multirow{2}{*}{Method} & \multicolumn{5}{c|}{Cartoon} & \multicolumn{5}{c|}{Game} & \multicolumn{5}{c}{Average} \\ \cline{2-16} 
     & \multicolumn{1}{c|}{AUC-J↑} & \multicolumn{1}{c|}{SIM↑} & \multicolumn{1}{c|}{s-AUC↑} & \multicolumn{1}{c|}{CC↑} & NSS↑ & \multicolumn{1}{c|}{AUC-J↑} & \multicolumn{1}{c|}{SIM↑} & \multicolumn{1}{c|}{s-AUC↑} & \multicolumn{1}{c|}{CC↑} & NSS↑ & \multicolumn{1}{c|}{AUC-J↑} & \multicolumn{1}{l|}{SIM↑} & \multicolumn{1}{l|}{s-AUC↑} & \multicolumn{1}{l|}{CC↑} & NSS↑ \\ \hline
    STSANet\cite{wang2021spatio} & 0.7476 & 0.3033 & 0.5481 & 0.3144 & 1.2218 & 0.8106 & 0.3582 & 0.5539 & 0.4300 & 1.6929 & 0.7786 & 0.3303 & 0.5509 & 0.3713 & 1.4536 \\
    HD2S\cite{bellitto2021hierarchical} & 0.8079 & 0.3670 & 0.5315 & 0.3897 & 1.5293 & 0.8149 & 0.4039 & 0.5283 & 0.4830 & 1.8880 & 0.8113 & 0.3851 & 0.5299 & 0.4356 & 1.7058 \\
    ViNet\cite{jain2021vinet} & 0.8258 & 0.3579 & 0.5716 & 0.3956 & 1.5458 & 0.8277 & 0.3811 & 0.5607 & 0.4527 & 1.7829 & 0.8267 & 0.3693 & 0.5662 & 0.4237 & 1.6625 \\
    UNISAL\cite{drostejiao2020} & 0.8194 & 0.3695 & 0.5653 & 0.3878 & 1.5334 & 0.8327 & 0.4109 & 0.5272 & 0.4105 & 1.8861 & 0.8259 & 0.3899 & 0.5466 & 0.3989 & 1.7069 \\
    TASED-Net\cite{min2019tased} & 0.8037 & 0.3643 & 0.5670 & 0.3876 & 1.5147 & 0.8165 & 0.3983 & 0.5603 & 0.4704 & 1.8700 & 0.8100 & 0.3810 & 0.5637 & 0.4283 & 1.6895 \\
    STAVis\cite{tsiami2020stavis} & 0.8055 & 0.3449 & 0.5489 & 0.3534 & 1.3535 & 0.8227 & 0.3824 & 0.5483 & 0.4542 & 1.7549 & 0.8140 & 0.3633 & 0.5486 & 0.4030 & 1.5510 \\
    STSANet*\cite{wang2021spatio} & 0.7910 & 0.2896 & 0.5277 & 0.3522 & 1.3431 & 0.8669 & 0.3520 & 0.5324 & 0.5857 & 2.3312 & 0.8283 & 0.3203 & 0.5300 & 0.4670 & 1.8292 \\
    HD2S*\cite{bellitto2021hierarchical} & 0.8298 & 0.3872 & 0.5199 & 0.4527 & 1.7395 & 0.8679 & 0.4601 & 0.5313 & 0.5979 & 2.3507 & 0.8485 & 0.4231 & 0.5255 & 0.5241 & 2.0402 \\
    ViNet*\cite{jain2021vinet} & 0.8443 & 0.3771 & 0.5726 & 0.4559 & 1.7573 & 0.8817 & 0.4464 & 0.5709 & 0.6034 & 2.4102 & 0.8627 & 0.4112 & 0.5718 & 0.5285 & 2.0785 \\
    UNISAL*\cite{drostejiao2020} & 0.8367 & 0.3811 & 0.5784 & 0.4417 & 1.6896 & 0.8713 & 0.4571 & 0.5323 & 0.5206 & 2.3414 & 0.8537 & 0.4185 & 0.5557 & 0.4805 & 2.0103 \\
    TASED-Net*\cite{min2019tased} & 0.8162 & 0.3695 & 0.5737 & 0.3947 & 1.5309 & 0.8600 & 0.4512 & 0.5708 & 0.5385 & 2.1869 & 0.8377 & 0.4097 & 0.5723 & 0.4654 & 1.8537 \\
    STAVis*\cite{tsiami2020stavis} & 0.8347 & 0.3838 & 0.5317 & 0.4408 & 1.6697 & 0.8674 & 0.4450 & 0.5343 & 0.5684 & 2.2028 & 0.8508 & 0.4139 & 0.5330 & 0.5036 & 1.9320 \\
    AViNet(v1)@\cite{jain2021vinet} & 0.8048 & 0.3332 & 0.5658 & 0.3566 & 1.4011 & 0.8187 & 0.3577 & 0.5549 & 0.4513 & 1.7803 & 0.8116 & 0.3453 & 0.5604 & 0.4032 & 1.5877 \\
    AViNet(v2)@\cite{jain2021vinet} & 0.7822 & 0.2964 & \textbf{0.5795} & 0.3102 & 1.1899 & 0.8130 & 0.3097 & \textbf{0.5751} & 0.3983 & 1.5463 & 0.7974 & 0.3029 & \textbf{0.5773} & 0.3535 & 1.3652 \\
    STAVis@\cite{tsiami2020stavis} & 0.8044 & 0.3425 & 0.5543 & 0.3513 & 1.3563 & 0.8222 & 0.3823 & 0.5491 & 0.4498 & 1.7758 & 0.8132 & 0.3621 & 0.5517 & 0.3998 & 1.5627 \\
    AViNet(v1)@*\cite{jain2021vinet} & 0.8463 & 0.3856 & 0.5705 & 0.4733 & 1.8314 & 0.8857 & 0.4585 & 0.5527 & 0.6193 & 2.5171 & 0.8657 & 0.4215 & 0.5617 & 0.5451 & 2.1688 \\
    AViNet(v2)@*\cite{jain2021vinet} & 0.8473 & 0.3821 & 0.5773 & 0.4869 & 1.8905 & 0.8861 & 0.4742 & 0.5677 & 0.6231 & 2.5688 & 0.8664 & 0.4274 & 0.5726 & 0.5539 & 2.2242 \\
    STAVis@*\cite{tsiami2020stavis} & 0.8325 & 0.3867 & 0.5443 & 0.4468 & 1.7127 & 0.8698 & 0.4459 & 0.5398 & 0.5702 & 2.2211 & 0.8509 & 0.4158 & 0.5421 & 0.5075 & 1.9628 \\
    Ours@ & \textbf{0.8537} & \textbf{0.3887} & 0.5618 & \textbf{0.4957} & \textbf{1.9192} & \textbf{0.8913} & \textbf{0.4743} & 0.5644 & \textbf{0.6347} & \textbf{2.5737} & \textbf{0.8722} & \textbf{0.4308} & 0.5631 & \textbf{0.5641} & \textbf{2.2412} \\ \hline
    \end{tabular}%
    }
    \label{tab:comparison}
    \vspace{-2mm}\end{table*}
    
\subsection{Implementation Details}
We utilize Pytorch \cite{paszke2019pytorch} to implement our code on two NVIDIA GeForce RTX 2080. As for hyperparameters, we train NPSNet using Adam \cite{kingma2014adam} with $\beta_{1}=0.9$ and $\beta_{2}=0.999$.
The training batch size is set to 8 and the learning rate is fixed to 0.0001. To save computing resources, we resize the input video frames from 640$\times$360 to 384$\times$224 during training and testing.
    
\subsection{Comparisons with SOTAs}
To demonstrate the effectiveness of NPSNet on non-photorealistic video, we conduct quantitative experiments with NPSNet and related methods on our NPF-200 dataset.
Cartoons and games are two different non-photorealistic domains in saliency detection. We thus train two types of visual-audio NPSNet models: 1) an NPSNet model trained on cartoon videos and 2) an NPSNet model trained on game videos.
These two models are compared with other visual-only and visual-audio methods on game and cartoon scenes, respectively (\ie, we use the model trained by game videos to test game scenes and use the model trained by cartoons to test cartoon scenes).

\textbf{Baselines.}
Many comparable methods are designed in a visual-only way. We compare our NPSNet model with six visual-only models: STSANet\cite{wang2021spatio}, HD2S\cite{bellitto2021hierarchical}, ViNet\cite{jain2021vinet}, UNISAL\cite{drostejiao2020}, TASED-Net\cite{min2019tased} and STAVis\cite{tsiami2020stavis}.
Besides, some methods take audio into account for assisting the process of saliency detection. We compare our NPSNet model with three visual-audio models: two model varients of AViNet\cite{jain2021vinet} and STAVis\cite{tsiami2020stavis}.
Note that we not only take these models from their original trained parameters on photorealistic scenes (DHF1K\cite{wang2018revisiting}) for evaluation, but we also retrain another variant on these models using our cartoon and game training videos.

\textbf{Evaluation metrics.}
We utilize five indicators commonly used in saliency detection for quantitative comparison \cite{borji2012state}, namely AUC-Judd (\textbf{AUC-J}), Similarity Metric (\textbf{SIM}), shuffled AUC (\textbf{s-AUC}), Linear Correlation Coefficient (\textbf{CC}) and Normalized Scanpath Saliency (\textbf{NSS}).
In addition to calculating these five indicators in the game and cartoon scenes, all models also calculate the average results by weighting the existing results according to the number of frames of game and cartoon videos in the test set.
    
\begin{table*}[t]
    \caption{Cross-domain analysis. We train models with data from different domains for testing cross-domain performance.}\vspace{-3mm}
    \label{tab:analysis}
    \resizebox{\textwidth}{!}{%
    \begin{tabular}{c|ccccc|ccccc|ccccc}
    \hline
    \multirow{2}{*}{Model} & \multicolumn{5}{c|}{DHF1K\cite{wang2018revisiting} testset} & \multicolumn{5}{c|}{Cartoon testset} & \multicolumn{5}{c}{Game testset} \\ \cline{2-16} 
     & \multicolumn{1}{c|}{AUC-J↑} & \multicolumn{1}{c|}{SIM↑} & \multicolumn{1}{c|}{s-AUC↑} & \multicolumn{1}{c|}{CC↑} & NSS↑ & \multicolumn{1}{c|}{AUC-J↑} & \multicolumn{1}{c|}{SIM↑} & \multicolumn{1}{c|}{s-AUC↑} & \multicolumn{1}{c|}{CC↑} & NSS↑ & \multicolumn{1}{c|}{AUC-J↑} & \multicolumn{1}{c|}{SIM↑} & \multicolumn{1}{c|}{s-AUC↑} & \multicolumn{1}{c|}{CC↑} & NSS↑ \\ \hline
    NPSNet-DHF1K & \textbf{0.9127} & \textbf{0.3793} & \textbf{0.7138} & \textbf{0.5205} & \textbf{2.9618} & 0.8136 & 0.3553 & \textbf{0.5679} & 0.3852 & 1.5182 & 0.8298 & 0.3971 & 0.5608 & 0.4828 & 1.9157 \\ \cline{1-1}
    NPSNet-cartoon & 0.8951 & 0.2820 & 0.6595 & 0.4369 & 2.4010 & \textbf{0.8537} & \textbf{0.3887} & 0.5618 & \textbf{0.4957} & \textbf{1.9392} & 0.8568 & 0.3904 & 0.5501 & 0.5615 & 2.1861 \\ \cline{1-1}
    NPSNet-game & 0.8868 & 0.2868 & 0.6411 & 0.4110 & 2.3069 & 0.8362 & 0.3774 & 0.5603 & 0.4696 & 1.8575 & \textbf{0.8913} & \textbf{0.4743} & \textbf{0.5644} & \textbf{0.6347} & \textbf{2.5737} \\ \hline
    \end{tabular}%
    }\vspace{-2mm}
    \end{table*}
    
\textbf{Results.}
The quantitative experimental results are shown in \reftab{comparison}. Note that the retrained models are marked with * and the models using audio are marked with @.
We can see that for the models trained in photorealistic scenes, their performance in cartoon and game videos is not as good as our NPSNet and their retrained ones in non-photorealistic scenes, which shows that the models trained in photorealistic scenes cannot directly apply to non-photorealistic scenarios (\eg, ``HD2S'' and ``Ours@'' model).
For models trained on cartoon and game respectively (mark with *), we can see that they achieve higher scores on multiple metrics on cartoon and game than models trained on photorealistic scenes (\eg, ``STSANet(*)''). Besides, for those models that incorporate audio, their results generally outperform models without sound (\eg, ``AViNet(v1)@*'' and ``ViNet*'').
These results show that our model can extract and combine audio and video features well, and achieve good saliency detection results and outperform other methods through inter-modal and intra-modal correspondences. It should be noted that the ``AVINet(v2)@*'' trained on the photorealistic scene outperforms its retrained ``AVINet(v2)@*'' and our model on the s-AUC metric on game and cartoon.
This is because s-AUC can indirectly describe whether the model is affected by center bias~\cite{bylinskii2018different}. For videos with the center-bias effect, non-fixation points are introduced in the center area to reduce the impact of center-bias when calculating s-AUC, but this process reduces the s-AUC scores.
In the non-photorealistic modal scene, the game and cartoon characters that people focus on are often deliberately set in the middle of the screen, while DHF1K's\cite{wang2018revisiting} videos come from daily scenes where people do not have a major focus.
Therefore, the NPSNet model trained by DHF1K is less affected by center bias and achieves higher s-AUC scores in game and cartoon scenes.
In conclusion, adding audio and retraining the model on game and cartoon respectively can raise the ceiling of saliency detection. Qualitative results are shown in Appendix.
        
    
\begin{figure}[t]
    \centering
    \captionsetup[subfloat]{labelformat=empty,justification=centering}
    \newcommand{\tmpwidth}{.31\linewidth}
    
    \subfloat[]{
        \begin{minipage}{.022\linewidth}
            \rotatebox{90}{  \ \ \ \ \ \ \ \footnotesize{previous frames}} 
            \rotatebox{90}{\ results}
            
    \end{minipage}}
    \subfloat[GT]{
        \begin{minipage}{\tmpwidth}
            \includegraphics[width=\linewidth]{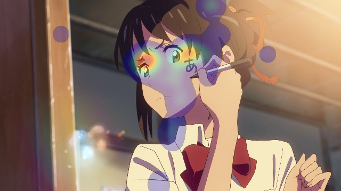}
            \includegraphics[width=\linewidth]{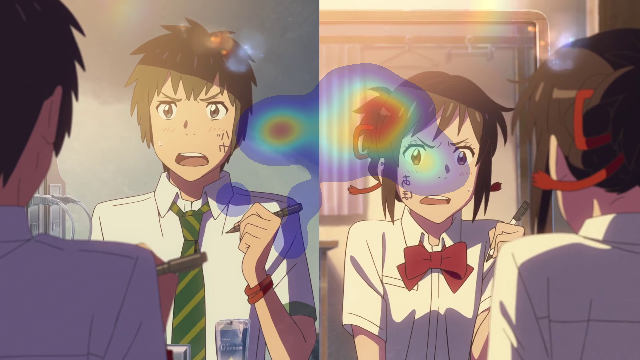}
    \end{minipage}}
    \subfloat[AViNet@*]{
        \begin{minipage}{\tmpwidth}
            \includegraphics[width=\linewidth]{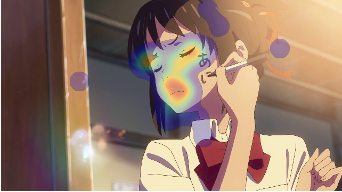}
            \includegraphics[width=\linewidth]{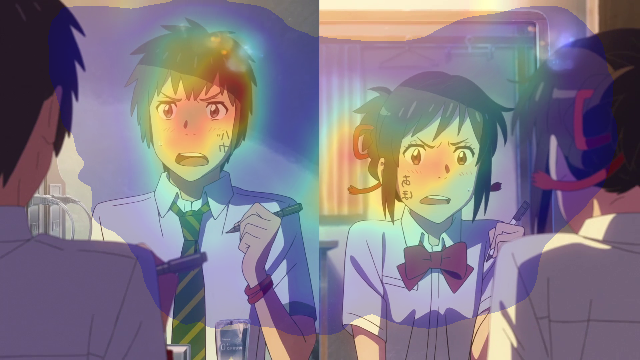}
    \end{minipage}}
    \subfloat[Ours@]{
        \begin{minipage}{\tmpwidth}
            \includegraphics[width=\linewidth]{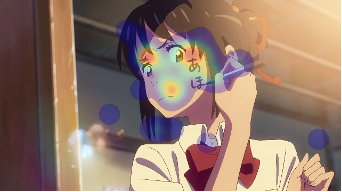}
            \includegraphics[width=\linewidth]{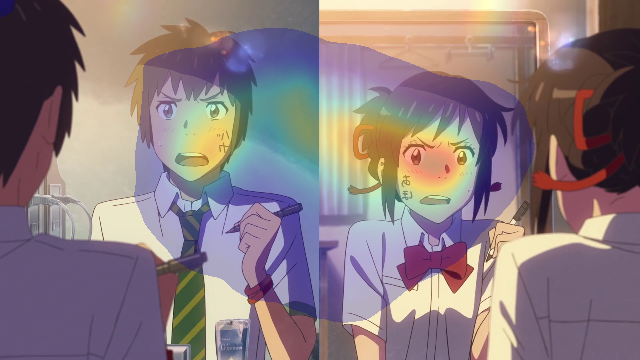}

    \end{minipage}}
    \vspace{-5mm}\caption{Example of long-term correlation. The girl's presence in previous frames leads to a higher attention bias toward her in the current frame.}
    \label{fig:temporal_compare}
    \vspace{-4mm}
\end{figure}

\subsection{Training Analysis}
\textbf{Cross-domain Training.} As our model contains multi-domain data, it naturally supports analysis for understanding model performance across different domains. We test on three domains: natural videos with DHF1K \cite{wang2018revisiting}, cartoon videos, and game videos with our dataset. We train our model on each of these three domains separately, obtaining three models, namely NPSNet-DHF1K, NPSNet-cartoon, and NPSNet-game, and test across them, with the results shown in \reftab{analysis}. We observe interesting findings from the results. As shown in the first row of \reftab{analysis}, the saliency detection model trained on the natural dataset (DHF1K) cannot be generalized directly to non-photorealistic (game and cartoon) domain scenes, with a larger performance difference. The model trained on a dataset under the same domain as the test one always performs the best. Although the s-AUC scores of NPSNet-DHF1K are similar to or even better than the other two models in the game and cartoon domain scenes, the underlying reason is that NPSNet-DHF1K is less affected by center-bias training videos. Furthermore, when NPSNet-cartoon is tested on the game domain, its performance does not drop significantly or even increase in some metrics, indicating that although game and cartoon belong to two domains, they have certain similarities.

\noindent\textbf{Long-term Correlation.} There are two main subjects in the current frame of \reffig{temporal_compare}, with the girl on the right appearing more frequently and occupying more screen space in the previous frames. This previous exposure to the girl influences viewers to pay more attention to her in the current frame, resulting in an uneven distribution of gaze points between the two subjects. Our method better captures the influence of long-term temporal information on saliency attention compared to previous methods and is closer to the ground truth.

\subsection{Multi-Modal Ablation Study}
In this section, we discuss the effectiveness of some key design factors of our model and build four model variants: 1) removing the inter-modal module and the audio branch (``w/o inter''); 2) removing the universal frequency-aware module (``w/o UFM''); 3) our full NPSNet (``full model'').
As shown in \reftab{ablation}, our full model achieves the best performance across all metrics, indicating that the audio modality can better guide visual multi-modal information at the feature level.
This enables audio to effectively assist the video for efficient and accurate saliency detection.

\begin{figure}[t]
    \centering
    \captionsetup[subfloat]{labelformat=empty,justification=centering}
    \newcommand{\tmpwidth}{.31\linewidth}
    
    \subfloat[]{
        \begin{minipage}{.022\linewidth}
            \rotatebox{90}{  \ \ \ \ \ \ \ \footnotesize{GT}} 
            \rotatebox{90}{\ w/ audio}
            \rotatebox{90}{w/o audio}
            
    \end{minipage}}
    \subfloat[]{
        \begin{minipage}{\tmpwidth}
            \includegraphics[width=\linewidth]{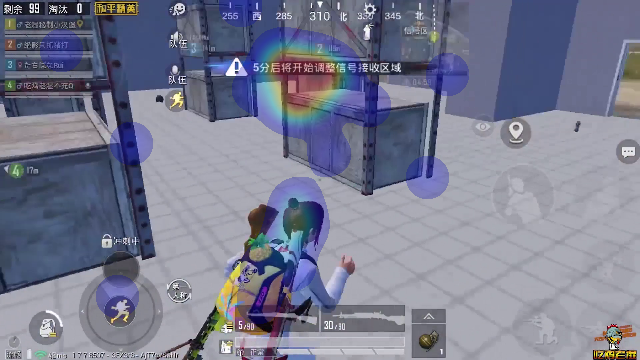}
            \includegraphics[width=\linewidth]{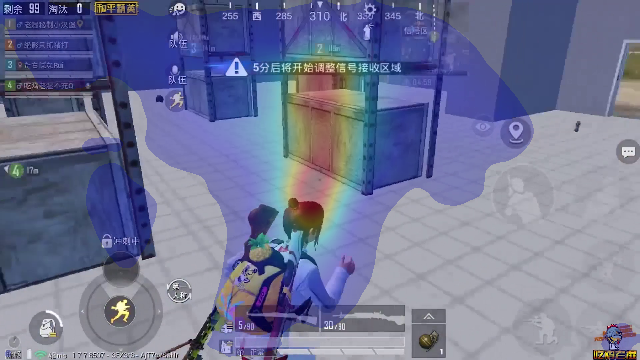}
            \includegraphics[width=\linewidth]{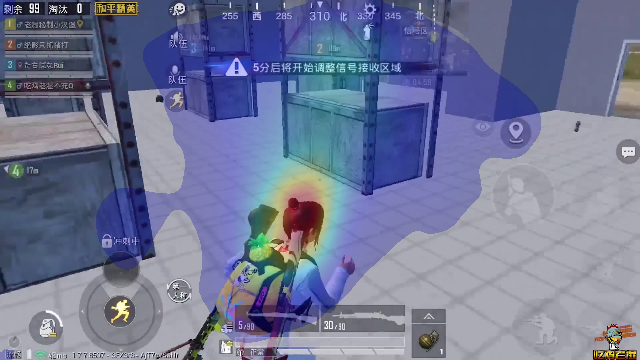}
    \end{minipage}}
    \subfloat[]{
        \begin{minipage}{\tmpwidth}
            \includegraphics[width=\linewidth]{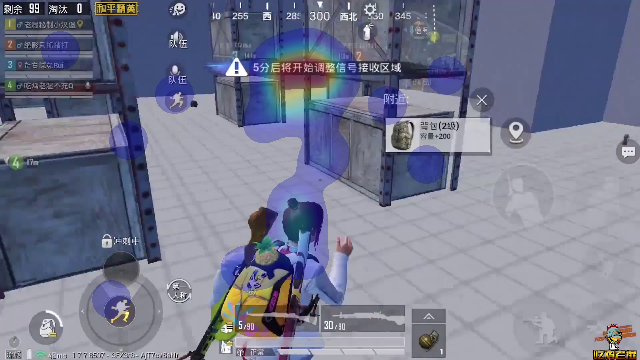}
            \includegraphics[width=\linewidth]{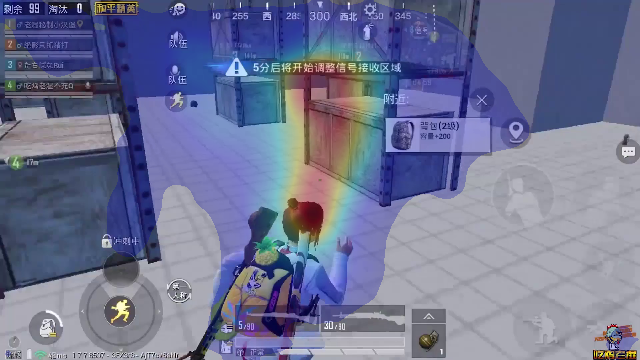}
            \includegraphics[width=\linewidth]{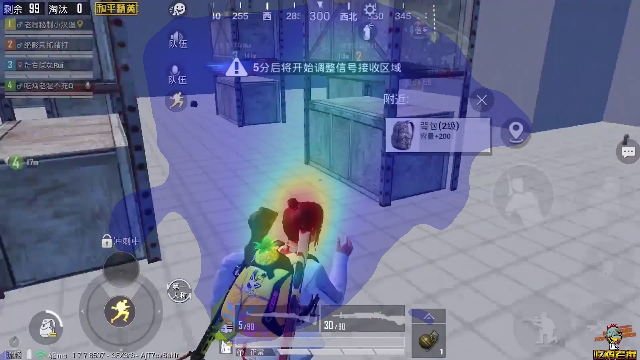}
    \end{minipage}}
    \subfloat[]{
        \begin{minipage}{\tmpwidth}
            \includegraphics[width=\linewidth]{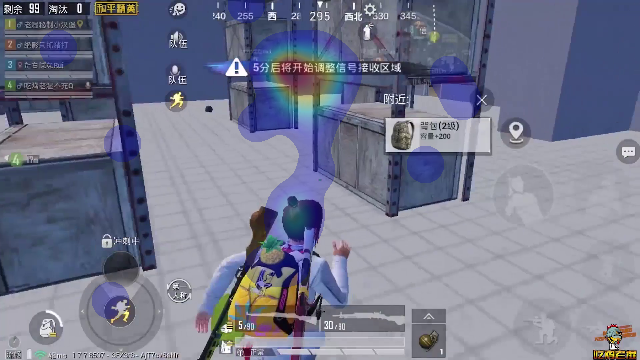}
            \includegraphics[width=\linewidth]{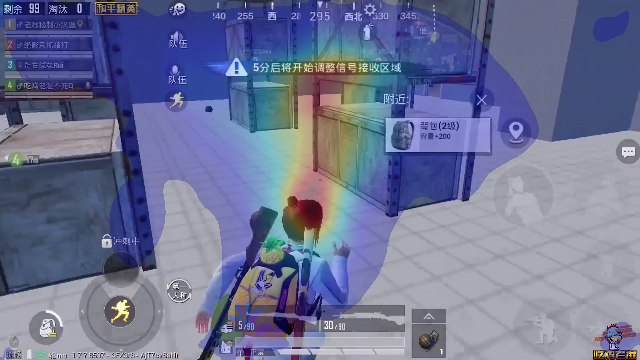}
            \includegraphics[width=\linewidth]{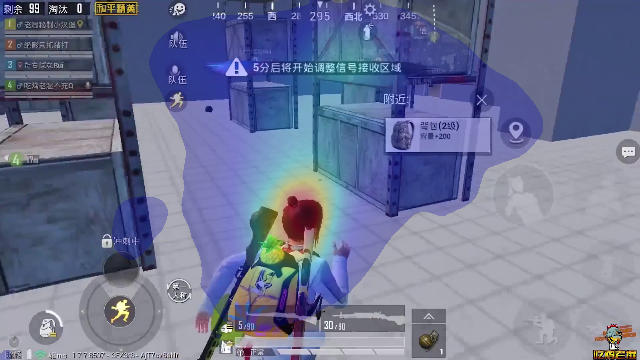}
    \end{minipage}}
    \vspace{-7mm}\caption{Comparison of saliency detection with and without audio input. The addition of audio provides supplementary information to guide the gaze toward key areas, while without audio, observers rely solely on image features.}\vspace{-5mm}
    \label{fig:visualization}
\end{figure}

To further explore the effectiveness of audio, we visualize the results of the audio-removed NPSNet (\ie, ``w/o inter'') and the full NPSNet, as shown in \reffig{visualization}.
When audio is absent, the saliency detection of the network relies solely on the features of video frames, and it is difficult for the network to focus on the observer's attention. However, after introducing audio, the saliency position of the network is obviously shifted to the area that the user pays attention to. This finding confirms that audio can provide useful cues for the network to learn saliency detection.

\begin{table}[t]
    \caption{
        Ablation study results.
    }\vspace{-3mm}
    \setlength\tabcolsep{0.5pt}
    \resizebox{\columnwidth}{!}{%
    \begin{tabular}{@{}c|ccccc|ccccc@{}}
    \toprule
    \multirow{2}{*}{Variants} & \multicolumn{5}{c|}{Cartoon} & \multicolumn{5}{c}{Game} \\ \cmidrule(l){2-11} 
     & \multicolumn{1}{c|}{AUC-J↑} & \multicolumn{1}{c|}{SIM↑} & \multicolumn{1}{c|}{s-AUC↑} & \multicolumn{1}{c|}{CC↑} & NSS↑ & \multicolumn{1}{c|}{AUC-J↑} & \multicolumn{1}{c|}{SIM↑} & \multicolumn{1}{c|}{s-AUC↑} & \multicolumn{1}{c|}{CC↑} & NSS↑ \\ \midrule
    w/o Inter & 0.8188 & 0.3556 & 0.5609 & 0.4092 & 1.6199 & 0.8593 & 0.4143 & 0.5607 & 0.5335 & 2.1425 \\ \cmidrule(r){1-1}
    w/o UFM & 0.8276 & 0.3677 & 0.5611 & 0.4201 & 1.7811 & 0.8671 & 0.4439 & 0.5601 & 0.5811 & 2.2231 \\ \cmidrule(r){1-1}
    Full Model & \textbf{0.8537} & \textbf{0.3887} & \textbf{0.5618} & \textbf{0.4957} & \textbf{1.9192} & \textbf{0.8913} & \textbf{0.4743} & \textbf{0.5644} & \textbf{0.6347} & \textbf{2.5737} \\ \bottomrule
    \end{tabular}%
    }
    \label{tab:ablation}
    \vspace{-4mm}\end{table}
    
\section{Conclusion}\label{conclusion}
In this paper, we present a new multi-modal non-photorealistic eye fixation video dataset and a multi-modal saliency detection network. Our dataset contains game and cartoon videos with rich semantic content and auxiliary audio, providing a benchmark for non-photorealistic video saliency detection. The proposed NPSNet extracts visual-audio features and builds a bridge for inter-modality interaction, achieving state-of-the-art performance on our dataset.
We have also conducted extensive experiments and analyses to gain insights into this new task. The results demonstrate the effectiveness of our model and dataset. We hope that our work will inspire further research in non-photorealistic video saliency detection and lead to new applications in areas such as editing, VR, and gaming.

\textbf{Acknowledgement.} The work is supported by the National Natural Science Foundation of China (No. 61972162); Guangdong Natural Science Funds for Distinguished Young Scholars (No. 2023B1515020097); Natural Science Foundation of Guangdong Province (2023A1515012894); Key R\&D Project of Guangzhou Science and Technology Plan (2023B01J0002); and Singapore Ministry of Education Academic Research Fund Tier 1 (MSS23C002).

{\small
\bibliographystyle{ieee_fullname}
\bibliography{egbib}

\begin{thebibliography}{10}\itemsep=-1pt

\bibitem{alkan2007studying}
Serkan Alkan and Kursat Cagiltay.
\newblock Studying computer game learning experience through eye tracking.
\newblock {\em British Journal of Educational Technology}, 38(3):538--542,
  2007.

\bibitem{aytar2016soundnet}
Yusuf Aytar, Carl Vondrick, and Antonio Torralba.
\newblock Soundnet: Learning sound representations from unlabeled video.
\newblock {\em NeurIPS}, 29, 2016.

\bibitem{bak2017spatio}
Cagdas Bak, Aysun Kocak, Erkut Erdem, and Aykut Erdem.
\newblock Spatio-temporal saliency networks for dynamic saliency prediction.
\newblock {\em IEEE Transactions on Multimedia}, 20(7):1688--1698, 2017.

\bibitem{bellitto2021hierarchical}
Giovanni Bellitto, Federica Proietto~Salanitri, Simone Palazzo, Francesco
  Rundo, Daniela Giordano, and Concetto Spampinato.
\newblock Hierarchical domain-adapted feature learning for video saliency
  prediction.
\newblock {\em International Journal of Computer Vision}, 129(12):3216--3232,
  2021.

\bibitem{borji2012state}
Ali Borji and Laurent Itti.
\newblock State-of-the-art in visual attention modeling.
\newblock {\em IEEE TPAMI}, 35(1):185--207, 2012.

\bibitem{borji2015cat2000}
Ali Borji and Laurent Itti.
\newblock Cat2000: A large scale fixation dataset for boosting saliency
  research.
\newblock {\em arXiv preprint arXiv:1505.03581}, 2015.

\bibitem{bruce2005saliency}
Neil Bruce and John Tsotsos.
\newblock Saliency based on information maximization.
\newblock {\em NeurIPS}, 18, 2005.

\bibitem{bullier2001integrated}
Jean Bullier.
\newblock Integrated model of visual processing.
\newblock {\em Brain research reviews}, 36(2-3):96--107, 2001.

\bibitem{bylinskii2018different}
Zoya Bylinskii, Tilke Judd, Aude Oliva, Antonio Torralba, and Fr{\'e}do Durand.
\newblock What do different evaluation metrics tell us about saliency models?
\newblock {\em IEEE TPAMI}, 41(3):740--757, 2018.

\bibitem{caffrey2008viewer}
Colm Caffrey.
\newblock Viewer perception of visual nonverbal cues in subtitled tv anime.
\newblock {\em European Journal of English Studies}, 12(2):163--178, 2008.

\bibitem{carion2020end}
Nicolas Carion, Francisco Massa, Gabriel Synnaeve, Nicolas Usunier, Alexander
  Kirillov, and Sergey Zagoruyko.
\newblock End-to-end object detection with transformers.
\newblock In {\em ECCV}, pages 213--229, 2020.

\bibitem{chang2021temporal}
Qinyao Chang and Shiping Zhu.
\newblock Temporal-spatial feature pyramid for video saliency detection.
\newblock {\em arXiv preprint arXiv:2105.04213}, 2021.

\bibitem{chen2022comprehensive}
Chenglizhao Chen, Mengke Song, Wenfeng Song, Li Guo, and Muwei Jian.
\newblock A comprehensive survey on video saliency detection with auditory
  information: the audio-visual consistency perceptual is the key!
\newblock {\em IEEE Transactions on Circuits and Systems for Video Technology},
  2022.

\bibitem{coutrot2014saliency}
Antoine Coutrot and Nathalie Guyader.
\newblock How saliency, faces, and sound influence gaze in dynamic social
  scenes.
\newblock {\em Journal of vision}, 14(8):5--5, 2014.

\bibitem{coutrot2015efficient}
Antoine Coutrot and Nathalie Guyader.
\newblock An efficient audiovisual saliency model to predict eye positions when
  looking at conversations.
\newblock In {\em European Signal Processing Conference}, pages 1531--1535,
  2015.

\bibitem{doherty2018development}
Stephen Doherty, Jan-Louis Kruger, T Dwyer, C Perkins, S Redmond, and J Sita.
\newblock The development of eye tracking in empirical research on subtitling
  and captioning.
\newblock {\em Seeing into screens: Eye tracking and the moving image}, pages
  46--64, 2018.

\bibitem{dosovitskiy2020image}
Alexey Dosovitskiy, Lucas Beyer, Alexander Kolesnikov, Dirk Weissenborn,
  Xiaohua Zhai, Thomas Unterthiner, Mostafa Dehghani, Matthias Minderer, Georg
  Heigold, Sylvain Gelly, et~al.
\newblock An image is worth 16x16 words: Transformers for image recognition at
  scale.
\newblock {\em arXiv preprint arXiv:2010.11929}, 2020.

\bibitem{droste2020unified}
Richard Droste, Jianbo Jiao, and J~Alison Noble.
\newblock Unified image and video saliency modeling.
\newblock In {\em ECCV}, pages 419--435, 2020.

\bibitem{drostejiao2020}
Richard {Droste}, Jianbo {Jiao}, and J.~Alison {Noble}.
\newblock {Unified Image and Video Saliency Modeling}.
\newblock In {\em ECCV}, 2020.

\bibitem{engelke2009visual}
Ulrich Engelke, Hans-Jurgen Zepernick, and Anthony Maeder.
\newblock Visual attention modeling: region-of-interest versus fixation
  patterns.
\newblock In {\em Picture Coding Symposium}, pages 1--4, 2009.

\bibitem{gollwitzer1999deliberative}
Peter~M Gollwitzer and Ute~C Bayer.
\newblock Deliberative versus implemental mindsets in the control of action,
  1999.

\bibitem{gygli2014creating}
Michael Gygli, Helmut Grabner, Hayko Riemenschneider, and Luc~Van Gool.
\newblock Creating summaries from user videos.
\newblock In {\em ECCV}, pages 505--520, 2014.

\bibitem{hadizadeh2011eye}
Hadi Hadizadeh, Mario~J Enriquez, and Ivan~V Bajic.
\newblock Eye-tracking database for a set of standard video sequences.
\newblock {\em IEEE TIP}, 21(2):898--903, 2011.

\bibitem{harel2006graph}
Jonathan Harel, Christof Koch, and Pietro Perona.
\newblock Graph-based visual saliency.
\newblock {\em NeurIPS}, 19, 2006.

\bibitem{he2019exploring}
Shengfeng He, Chu Han, Guoqiang Han, and Jing Qin.
\newblock Exploring duality in visual question-driven top-down saliency.
\newblock {\em IEEE transactions on neural networks and learning systems},
  31(7):2672--2679, 2019.

\bibitem{he2017delving}
Shengfeng He, Jianbo Jiao, Xiaodan Zhang, Guoqiang Han, and Rynson~WH Lau.
\newblock Delving into salient object subitizing and detection.
\newblock In {\em Proceedings of the IEEE International Conference on Computer
  Vision}, pages 1059--1067, 2017.

\bibitem{he2014saliency}
Shengfeng He and Rynson~WH Lau.
\newblock Saliency detection with flash and no-flash image pairs.
\newblock In {\em Computer Vision--ECCV 2014: 13th European Conference, Zurich,
  Switzerland, September 6-12, 2014, Proceedings, Part III 13}, pages 110--124.
  Springer, 2014.

\bibitem{he2016exemplar}
Shengfeng He, Rynson~WH Lau, and Qingxiong Yang.
\newblock Exemplar-driven top-down saliency detection via deep association.
\newblock In {\em Proceedings of the IEEE Conference on Computer Vision and
  Pattern Recognition}, pages 5723--5732, 2016.

\bibitem{jain2021vinet}
Samyak Jain, Pradeep Yarlagadda, Shreyank Jyoti, Shyamgopal Karthik, Ramanathan
  Subramanian, and Vineet Gandhi.
\newblock Vinet: Pushing the limits of visual modality for audio-visual
  saliency prediction.
\newblock In {\em 2021 IEEE/RSJ International Conference on Intelligent Robots
  and Systems (IROS)}, pages 3520--3527, 2021.

\bibitem{jiang2018deepvs}
Lai Jiang, Mai Xu, Tie Liu, Minglang Qiao, and Zulin Wang.
\newblock Deepvs: A deep learning based video saliency prediction approach.
\newblock In {\em Proceedings of the european conference on computer vision
  (eccv)}, pages 602--617, 2018.

\bibitem{jiang2017exploiting}
Yu-Gang Jiang, Zuxuan Wu, Jun Wang, Xiangyang Xue, and Shih-Fu Chang.
\newblock Exploiting feature and class relationships in video categorization
  with regularized deep neural networks.
\newblock {\em IEEE TPAMI}, 40(2):352--364, 2017.

\bibitem{jie2008video}
Li Jie and James~J Clark.
\newblock Video game design using an eye-movement-dependent model of visual
  attention.
\newblock {\em ACM Transactions on Multimedia Computing, Communications, and
  Applications}, 4(3):1--16, 2008.

\bibitem{kay2017kinetics}
Will Kay, Joao Carreira, Karen Simonyan, Brian Zhang, Chloe Hillier, Sudheendra
  Vijayanarasimhan, Fabio Viola, Tim Green, Trevor Back, Paul Natsev, et~al.
\newblock The kinetics human action video dataset.
\newblock {\em arXiv preprint arXiv:1705.06950}, 2017.

\bibitem{kenny2005preliminary}
Alan Kenny, Hendrik Koesling, Declan Delaney, Seamus McLoone, and Tomas~E Ward.
\newblock A preliminary investigation into eye gaze data in a first person
  shooter game.
\newblock 2005.

\bibitem{kingma2014adam}
Diederik~P Kingma and Jimmy Ba.
\newblock Adam: A method for stochastic optimization.
\newblock In {\em ICLR}, 2015.

\bibitem{koutras2015perceptually}
Petros Koutras and Petros Maragos.
\newblock A perceptually based spatio-temporal computational framework for
  visual saliency estimation.
\newblock {\em Signal Processing: Image Communication}, 38:15--31, 2015.

\bibitem{koutras2019susinet}
Petros Koutras and Petros Maragos.
\newblock Susinet: See, understand and summarize it.
\newblock In {\em CVPRW}, 2019.

\bibitem{lai2019video}
Qiuxia Lai, Wenguan Wang, Hanqiu Sun, and Jianbing Shen.
\newblock Video saliency prediction using spatiotemporal residual attentive
  networks.
\newblock {\em IEEE TIP}, 29:1113--1126, 2019.

\bibitem{le2006coherent}
Olivier Le~Meur, Patrick Le~Callet, Dominique Barba, and Dominique Thoreau.
\newblock A coherent computational approach to model bottom-up visual
  attention.
\newblock {\em IEEE TPAMI}, 28(5):802--817, 2006.

\bibitem{liu2022video}
Ze Liu, Jia Ning, Yue Cao, Yixuan Wei, Zheng Zhang, Stephen Lin, and Han Hu.
\newblock Video swin transformer.
\newblock In {\em CVPR}, pages 3202--3211, 2022.

\bibitem{ma2022video}
Cheng Ma, Haowen Sun, Yongming Rao, Jie Zhou, and Jiwen Lu.
\newblock Video saliency forecasting transformer.
\newblock {\em IEEE TCSVT}, 2022.

\bibitem{mathe2014actions}
Stefan Mathe and Cristian Sminchisescu.
\newblock Actions in the eye: Dynamic gaze datasets and learnt saliency models
  for visual recognition.
\newblock {\em IEEE transactions on pattern analysis and machine intelligence},
  37(7):1408--1424, 2014.

\bibitem{min2019tased}
Kyle Min and Jason~J Corso.
\newblock Tased-net: Temporally-aggregating spatial encoder-decoder network for
  video saliency detection.
\newblock In {\em ICCV}, pages 2394--2403, 2019.

\bibitem{min2014sound}
Xiongkuo Min, Guangtao Zhai, Zhongpai Gao, Chunjia Hu, and Xiaokang Yang.
\newblock Sound influences visual attention discriminately in videos.
\newblock In {\em 2014 Sixth International Workshop on Quality of Multimedia
  Experience (QoMEX)}, pages 153--158, 2014.

\bibitem{min2016fixation}
Xiongkuo Min, Guangtao Zhai, Ke Gu, and Xiaokang Yang.
\newblock Fixation prediction through multimodal analysis.
\newblock {\em ACM Transactions on Multimedia Computing, Communications, and
  Applications}, 13(1):1--23, 2016.

\bibitem{min2020multimodal}
Xiongkuo Min, Guangtao Zhai, Jiantao Zhou, Xiao-Ping Zhang, Xiaokang Yang, and
  Xinping Guan.
\newblock A multimodal saliency model for videos with high audio-visual
  correspondence.
\newblock {\em IEEE Transactions on Image Processing}, 29:3805--3819, 2020.

\bibitem{mital2011clustering}
Parag~K Mital, Tim~J Smith, Robin~L Hill, and John~M Henderson.
\newblock Clustering of gaze during dynamic scene viewing is predicted by
  motion.
\newblock {\em Cognitive computation}, 3(1):5--24, 2011.

\bibitem{nguyen2013static}
Tam~V Nguyen, Mengdi Xu, Guangyu Gao, Mohan Kankanhalli, Qi Tian, and Shuicheng
  Yan.
\newblock Static saliency vs. dynamic saliency: a comparative study.
\newblock In {\em Proceedings of the 21st ACM international conference on
  Multimedia}, pages 987--996, 2013.

\bibitem{niu2020boundary}
Yuzhen Niu, Guanchao Long, Wenxi Liu, Wenzhong Guo, and Shengfeng He.
\newblock Boundary-aware rgbd salient object detection with cross-modal feature
  sampling.
\newblock {\em IEEE Transactions on Image Processing}, 29:9496--9507, 2020.

\bibitem{park2022vision}
Namuk Park and Songkuk Kim.
\newblock How do vision transformers work?
\newblock {\em arXiv preprint arXiv:2202.06709}, 2022.

\bibitem{paszke2019pytorch}
Adam Paszke, Sam Gross, Francisco Massa, Adam Lerer, James Bradbury, Gregory
  Chanan, Trevor Killeen, Zeming Lin, Natalia Gimelshein, Luca Antiga, et~al.
\newblock Pytorch: An imperative style, high-performance deep learning library.
\newblock {\em NeurIPS}, 32, 2019.

\bibitem{perrott1990auditory}
David~R Perrott, Kourosh Saberi, Kathleen Brown, and Thomas~Z Strybel.
\newblock Auditory psychomotor coordination and visual search performance.
\newblock {\em Perception \& psychophysics}, 48(3):214--226, 1990.

\bibitem{ren2020tenet}
Sucheng Ren, Chu Han, Xin Yang, Guoqiang Han, and Shengfeng He.
\newblock Tenet: Triple excitation network for video salient object detection.
\newblock In {\em Computer Vision--ECCV 2020: 16th European Conference,
  Glasgow, UK, August 23--28, 2020, Proceedings, Part V 16}, pages 212--228.
  Springer, 2020.

\bibitem{ren2021reciprocal}
Sucheng Ren, Wenxi Liu, Yongtuo Liu, Haoxin Chen, Guoqiang Han, and Shengfeng
  He.
\newblock Reciprocal transformations for unsupervised video object
  segmentation.
\newblock In {\em Proceedings of the IEEE/CVF conference on computer vision and
  pattern recognition}, pages 15455--15464, 2021.

\bibitem{rodriguez2008action}
Mikel~D Rodriguez, Javed Ahmed, and Mubarak Shah.
\newblock Action mach a spatio-temporal maximum average correlation height
  filter for action recognition.
\newblock In {\em CVPR}, pages 1--8, 2008.

\bibitem{saaty2022game}
Morva Saaty and Mahmoud~Reza Hashemi.
\newblock Game audio impacts on players’ visual attention, model performance
  for cloud gaming.
\newblock In {\em Symposium on Eye Tracking Research and Applications}, pages
  1--7, 2022.

\bibitem{seif2004visually}
Magy Seif El-Nasr and Chinmay Rao.
\newblock Visually directing user’s attention in interactive 3d environments.
\newblock 2004.

\bibitem{song2013different}
Guanghan Song, Denis Pellerin, and Lionel Granjon.
\newblock Different types of sounds influence gaze differently in videos.
\newblock {\em Journal of Eye Movement Research}, 6(4):1--13, 2013.

\bibitem{sultani2014human}
Waqas Sultani and Imran Saleemi.
\newblock Human action recognition across datasets by foreground-weighted
  histogram decomposition.
\newblock In {\em CVPR}, pages 764--771, 2014.

\bibitem{sundstedt2008psychophysical}
Veronica Sundstedt, Efstathios Stavrakis, Michael Wimmer, and Erik Reinhard.
\newblock A psychophysical study of fixation behavior in a computer game.
\newblock In {\em Proceedings of the symposium on Applied perception in
  graphics and visualization}, pages 43--50, 2008.

\bibitem{tavakoli2019dave}
Hamed~R Tavakoli, Ali Borji, Esa Rahtu, and Juho Kannala.
\newblock Dave: A deep audio-visual embedding for dynamic saliency prediction.
\newblock {\em arXiv preprint arXiv:1905.10693}, 2019.

\bibitem{tsiami2019behaviorally}
Antigoni Tsiami, Petros Koutras, Athanasios Katsamanis, Argiro Vatakis, and
  Petros Maragos.
\newblock A behaviorally inspired fusion approach for computational audiovisual
  saliency modeling.
\newblock {\em Signal Processing: Image Communication}, 76:186--200, 2019.

\bibitem{tsiami2020stavis}
Antigoni Tsiami, Petros Koutras, and Petros Maragos.
\newblock Stavis: Spatio-temporal audiovisual saliency network.
\newblock In {\em CVPR}, pages 4766--4776, 2020.

\bibitem{vaswani2017attention}
Ashish Vaswani, Noam Shazeer, Niki Parmar, Jakob Uszkoreit, Llion Jones,
  Aidan~N Gomez, {\L}ukasz Kaiser, and Illia Polosukhin.
\newblock Attention is all you need.
\newblock {\em NeurIPS}, 30, 2017.

\bibitem{vroomen2000sound}
Jean Vroomen and Beatrice~de Gelder.
\newblock Sound enhances visual perception: cross-modal effects of auditory
  organization on vision.
\newblock {\em Journal of experimental psychology: Human perception and
  performance}, 26(5):1583, 2000.

\bibitem{wang2020learning}
Bo Wang, Wenxi Liu, Guoqiang Han, and Shengfeng He.
\newblock Learning long-term structural dependencies for video salient object
  detection.
\newblock {\em IEEE Transactions on Image Processing}, 29:9017--9031, 2020.

\bibitem{wang2021semantic}
Guotao Wang, Chenglizhao Chen, Deng-Ping Fan, Aimin Hao, and Hong Qin.
\newblock From semantic categories to fixations: A novel weakly-supervised
  visual-auditory saliency detection approach.
\newblock In {\em Proceedings of the IEEE/CVF conference on computer vision and
  pattern recognition}, pages 15119--15128, 2021.

\bibitem{wang2018revisiting}
Wenguan Wang, Jianbing Shen, Fang Guo, Ming-Ming Cheng, and Ali Borji.
\newblock Revisiting video saliency: A large-scale benchmark and a new model.
\newblock In {\em CVPR}, pages 4894--4903, 2018.

\bibitem{wang2019revisiting}
Wenguan Wang, Jianbing Shen, Jianwen Xie, Ming-Ming Cheng, Haibin Ling, and Ali
  Borji.
\newblock Revisiting video saliency prediction in the deep learning era.
\newblock {\em IEEE transactions on pattern analysis and machine intelligence},
  43(1):220--237, 2019.

\bibitem{wang2021spatio}
Ziqiang Wang, Zhi Liu, Gongyang Li, Yang Wang, Tianhong Zhang, Lihua Xu, and
  Jijun Wang.
\newblock Spatio-temporal self-attention network for video saliency prediction.
\newblock {\em IEEE Transactions on Multimedia}, 2021.

\bibitem{wu2020salsac}
Xinyi Wu, Zhenyao Wu, Jinglin Zhang, Lili Ju, and Song Wang.
\newblock Salsac: A video saliency prediction model with shuffled attentions
  and correlation-based convlstm.
\newblock In {\em AAAI}, volume~34, pages 12410--12417, 2020.

\bibitem{xie2018rethinking}
Saining Xie, Chen Sun, Jonathan Huang, Zhuowen Tu, and Kevin Murphy.
\newblock Rethinking spatiotemporal feature learning: Speed-accuracy trade-offs
  in video classification.
\newblock In {\em Proceedings of the European conference on computer vision
  (ECCV)}, pages 305--321, 2018.

\bibitem{zhang2018video}
Kao Zhang and Zhenzhong Chen.
\newblock Video saliency prediction based on spatial-temporal two-stream
  network.
\newblock {\em IEEE TCSVT}, 29(12):3544--3557, 2018.

\bibitem{zhang2008sun}
Lingyun Zhang, Matthew~H Tong, Tim~K Marks, Honghao Shan, and Garrison~W
  Cottrell.
\newblock Sun: A bayesian framework for saliency using natural statistics.
\newblock {\em Journal of vision}, 8(7):32--32, 2008.

\bibitem{zhang2020atari}
Ruohan Zhang, Calen Walshe, Zhuode Liu, Lin Guan, Karl Muller, Jake Whritner,
  Luxin Zhang, Mary Hayhoe, and Dana Ballard.
\newblock Atari-head: Atari human eye-tracking and demonstration dataset.
\newblock In {\em Proceedings of the AAAI conference on artificial
  intelligence}, volume~34, pages 6811--6820, 2020.

\end{thebibliography}
}

\end{document}


\title[NPF-200: A Multi-Modal Eye Fixation Dataset and Method for Non-Photorealistic Videos]{NPF-200: A Multi-Modal Eye Fixation Dataset and Method\\ for Non-Photorealistic {Videos}\\--Appendix--}

\author{Ziyu Yang}
\authornote{Both authors contributed equally to this research.}
\affiliation{%
  \institution{South China University of Technology}
}
\email{yangzy.dlut@gmail.com}

\author{Sucheng Ren}
\authornotemark[1]
\author{Zongwei Wu}
\affiliation{%
	\institution{Singapore Management University}
}
\email{
oliverrensu@gmail.com
}
\email{zongweiwu999@gmail.com
}

\author{Nanxuan Zhao}
\affiliation{%
  \institution{Adobe Research}
}
\email{nanxuanzhao@gmail.com
}

\author{Junle Wang}
\affiliation{%
  \institution{Tencent}
}
\email{jljunlewang@tencent.com
}

\author{Jing Qin}
\affiliation{%
  \institution{The Hong Kong Polytechnic University}
}
\email{harry.qin@polyu.edu.hk}

\author{Shengfeng He}
\orcid{0000-0002-3802-4644}
\authornote{Corresponding author.}
\affiliation{%
	\institution{Singapore Management University}
}
\email{shengfenghe@smu.edu.sg}

\renewcommand{\shortauthors}{Yang et al.}

\begin{teaserfigure}
        \centering
        \captionsetup[subfloat]{labelformat=empty,justification=centering}
        \newcommand{\tmpwidth}{.116\linewidth}
        \subfloat[HD2S*]{
            \begin{minipage}{\tmpwidth}
                \includegraphics[width=\linewidth]{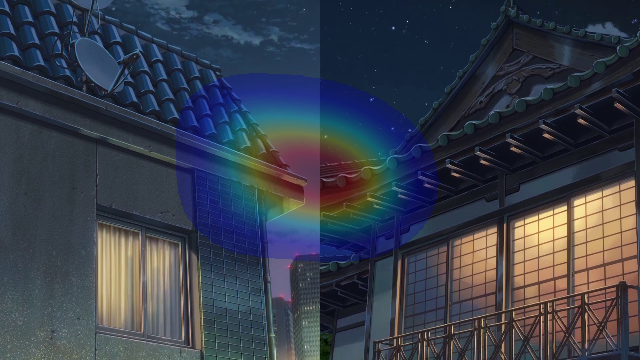}\\
                \includegraphics[width=\linewidth]{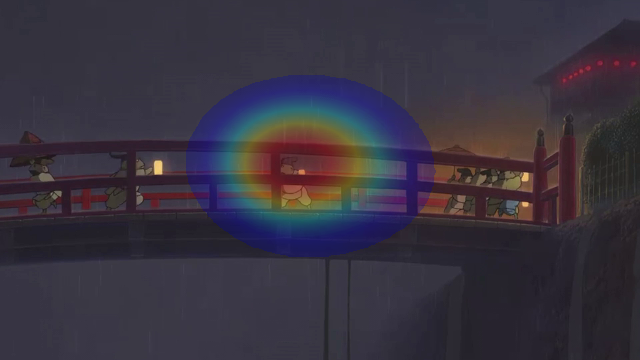}\\
                \includegraphics[width=\linewidth]{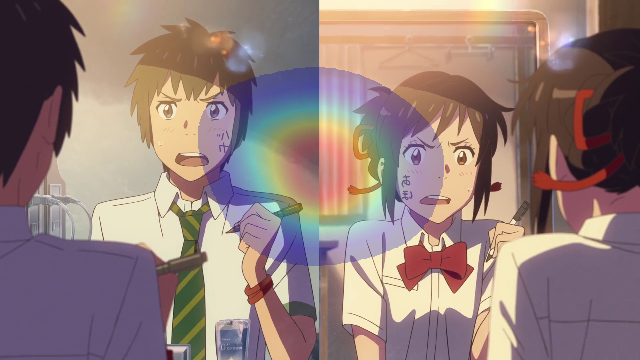}\\
                \includegraphics[width=\linewidth]{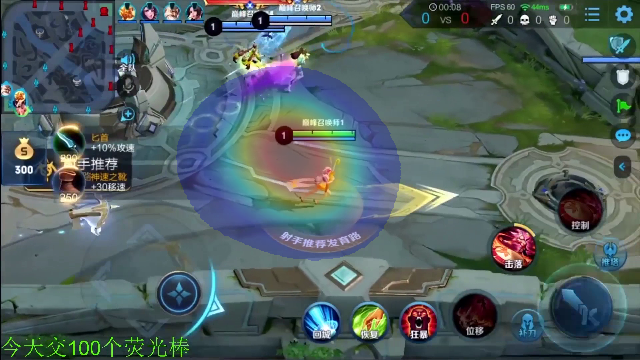}\\
                \includegraphics[width=\linewidth]{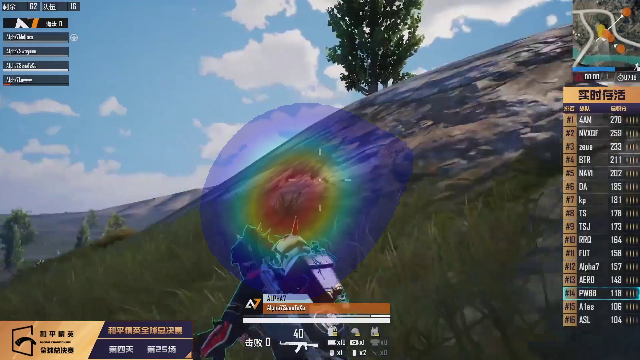}
            \end{minipage}
        }
        \subfloat[ViNet*]{
            \begin{minipage}{\tmpwidth}
                \includegraphics[width=\linewidth]{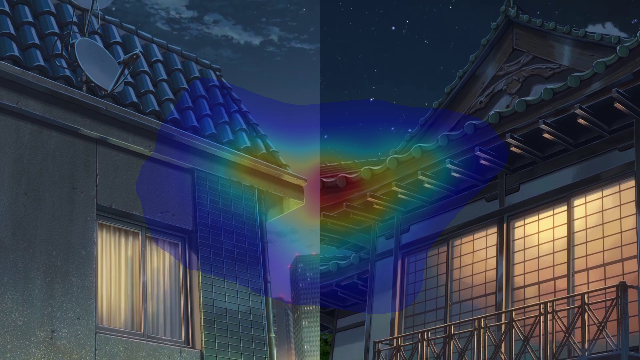}\\
                \includegraphics[width=\linewidth]{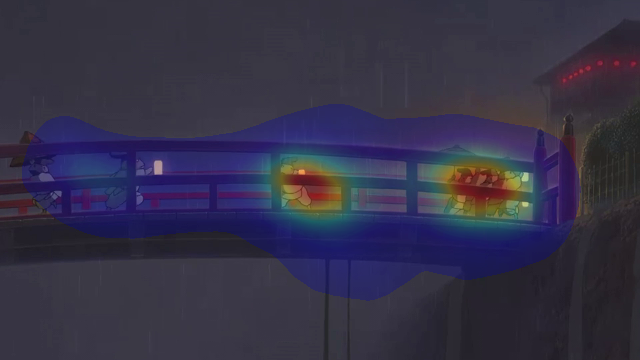}\\
                \includegraphics[width=\linewidth]{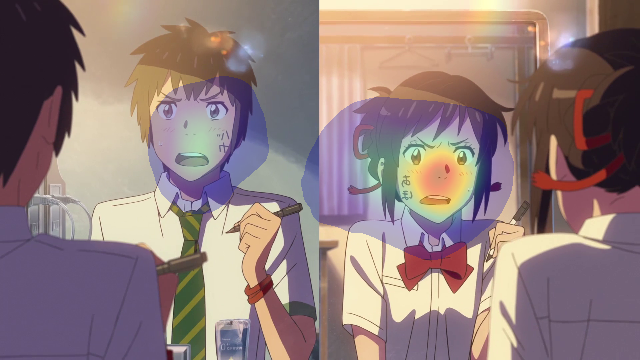}\\
                \includegraphics[width=\linewidth]{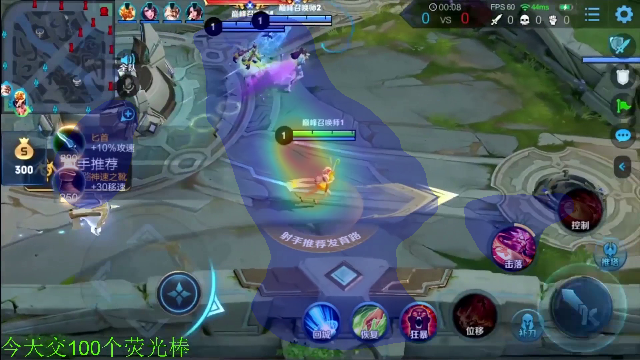}\\
                \includegraphics[width=\linewidth]{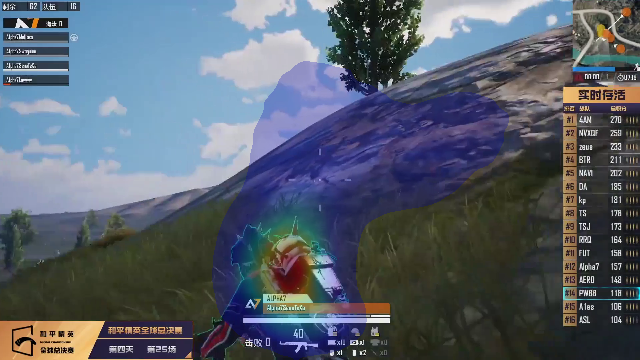}
            \end{minipage}
        }
        \subfloat[UNISAL*]{
            \begin{minipage}{\tmpwidth}
                \includegraphics[width=\linewidth]{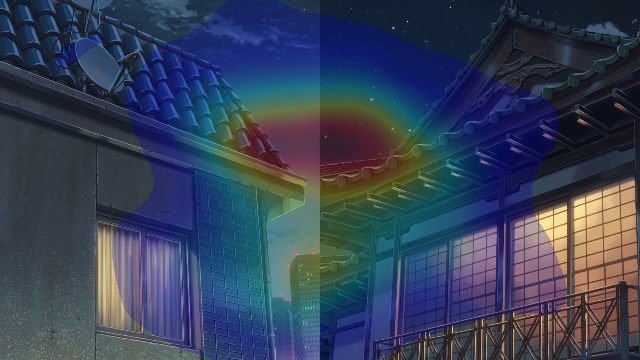}\\
                \includegraphics[width=\linewidth]{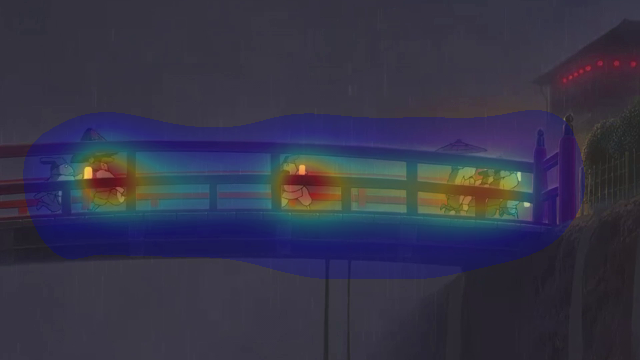}\\
                \includegraphics[width=\linewidth]{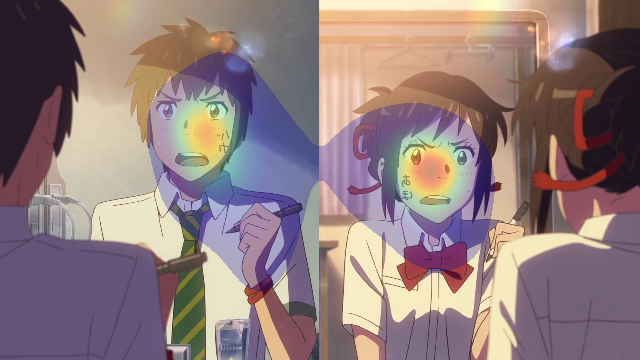}\\
                \includegraphics[width=\linewidth]{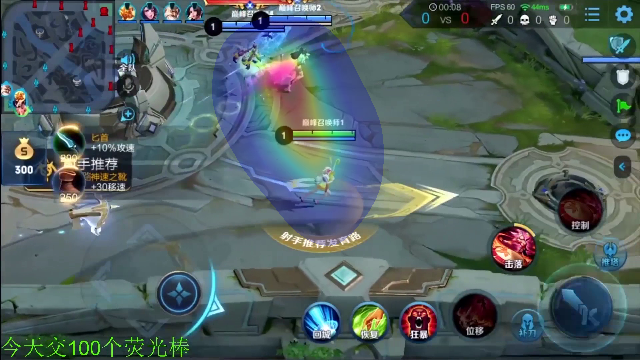}\\
                \includegraphics[width=\linewidth]{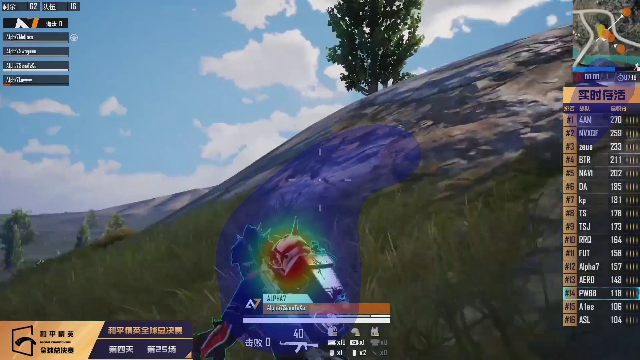}
            \end{minipage}
        }
        \subfloat[TASED-Net*]{
            \begin{minipage}{\tmpwidth}
                \includegraphics[width=\linewidth]{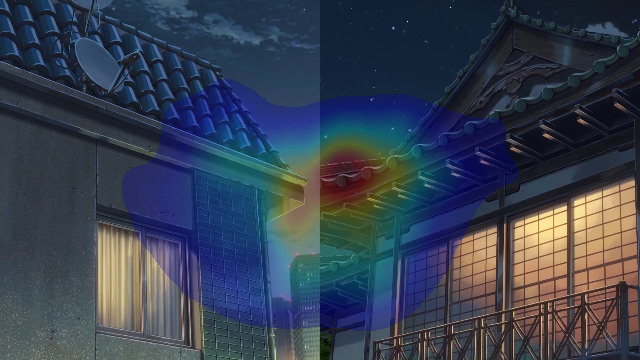}\\
                \includegraphics[width=\linewidth]{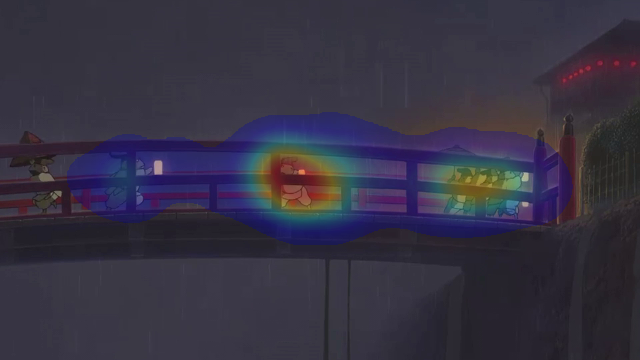}\\
                \includegraphics[width=\linewidth]{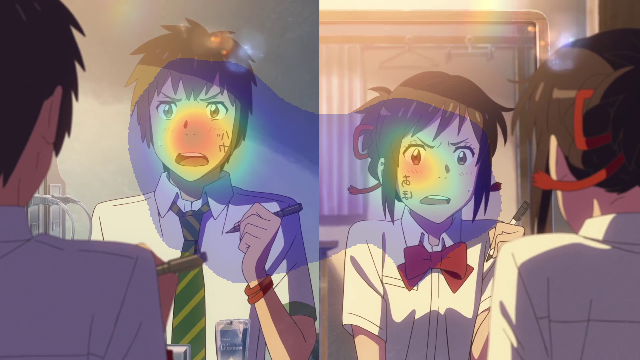}\\
                \includegraphics[width=\linewidth]{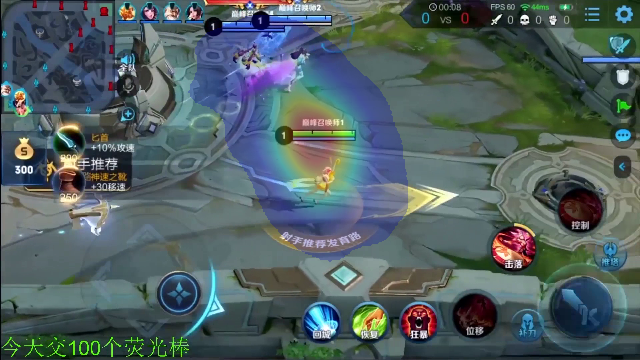}\\
                \includegraphics[width=\linewidth]{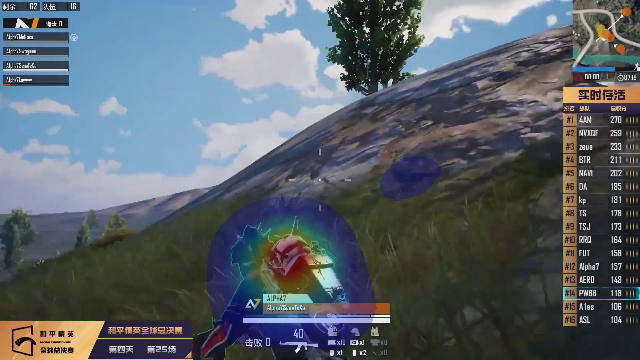}
            \end{minipage}
        }
        \subfloat[AViNet@*]{
            \begin{minipage}{\tmpwidth}
                \includegraphics[width=\linewidth]{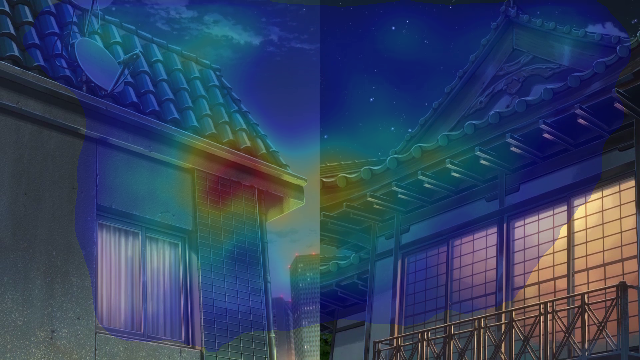}\\
                \includegraphics[width=\linewidth]{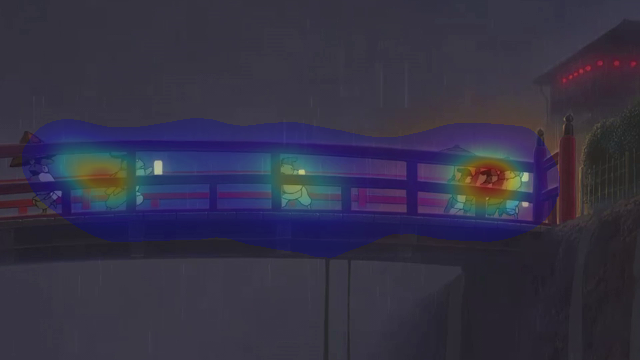}\\
                \includegraphics[width=\linewidth]{figure/experiment/results/3/avinet.png}\\
                \includegraphics[width=\linewidth]{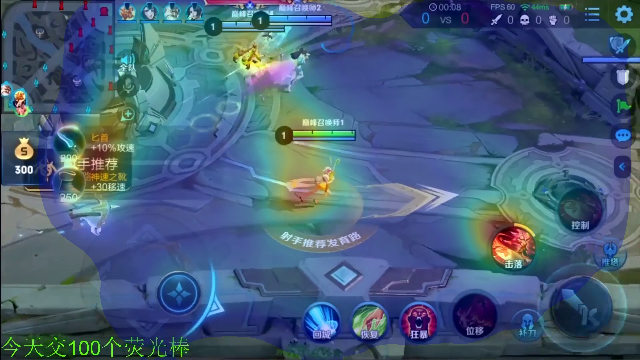}\\
                \includegraphics[width=\linewidth]{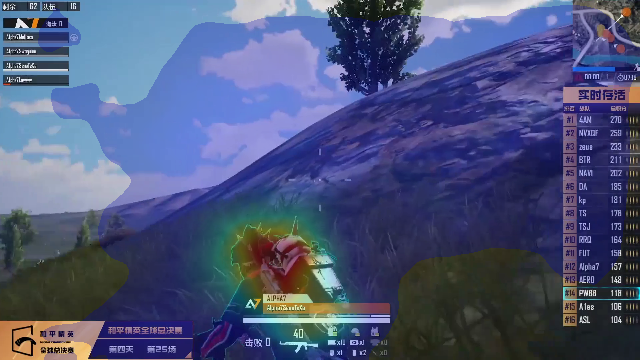}
            \end{minipage}
        }
        \subfloat[STAVis@*]{
            \begin{minipage}{\tmpwidth}
                \includegraphics[width=\linewidth]{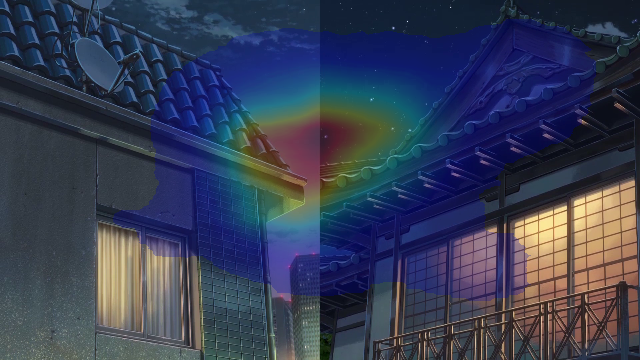}\\
                \includegraphics[width=\linewidth]{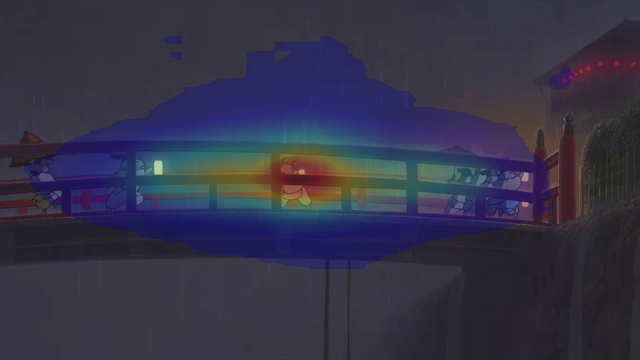}\\
                \includegraphics[width=\linewidth]{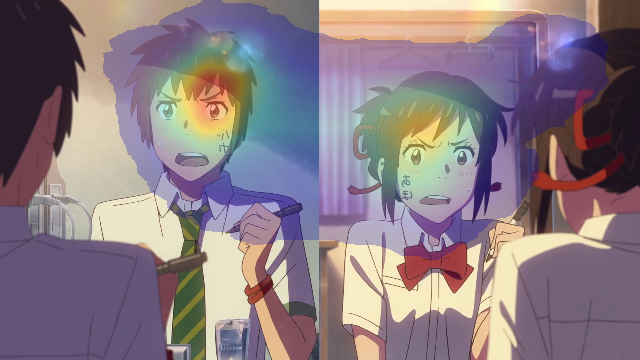}\\
                \includegraphics[width=\linewidth]{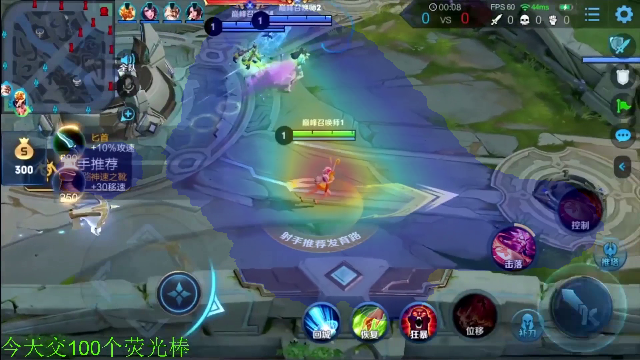}\\
                \includegraphics[width=\linewidth]{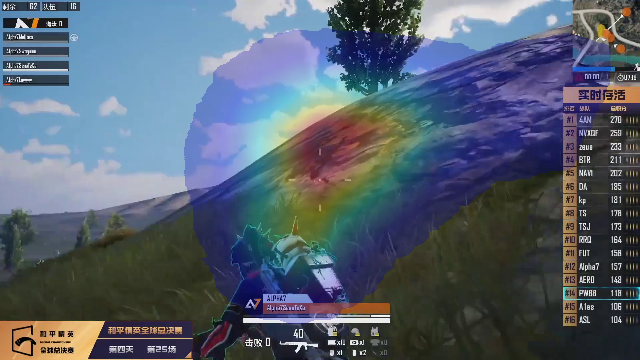}
            \end{minipage}
        }
        \subfloat[Ours@]{
            \begin{minipage}{\tmpwidth}
                \includegraphics[width=\linewidth]{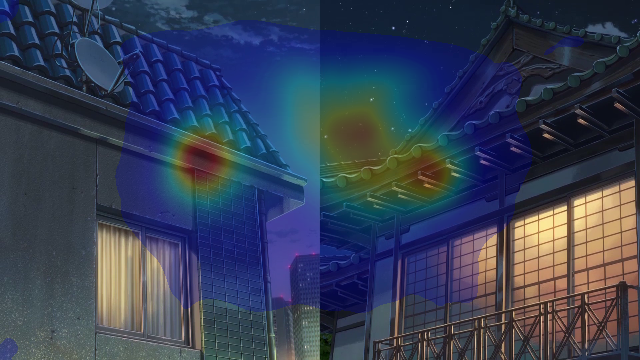}\\
                \includegraphics[width=\linewidth]{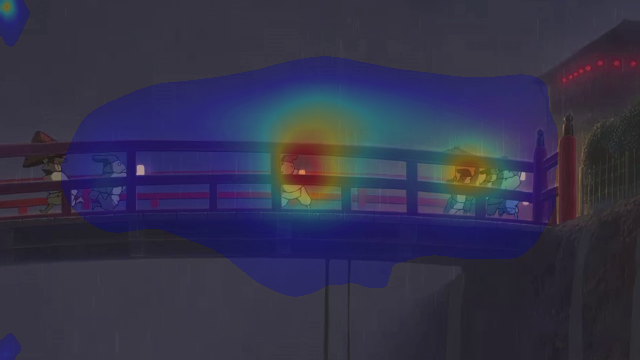}\\
                \includegraphics[width=\linewidth]{figure/experiment/results/3/NPS.png}\\
                \includegraphics[width=\linewidth]{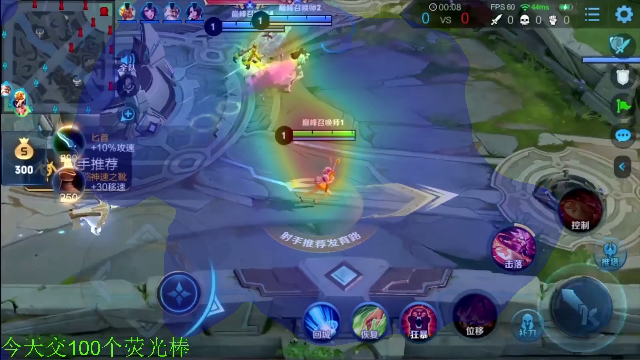}\\
                \includegraphics[width=\linewidth]{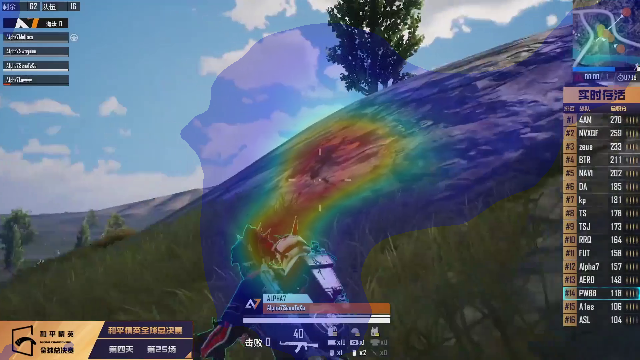}
            \end{minipage}
        }
        \subfloat[GT]{
            \begin{minipage}{\tmpwidth}
                \includegraphics[width=\linewidth]{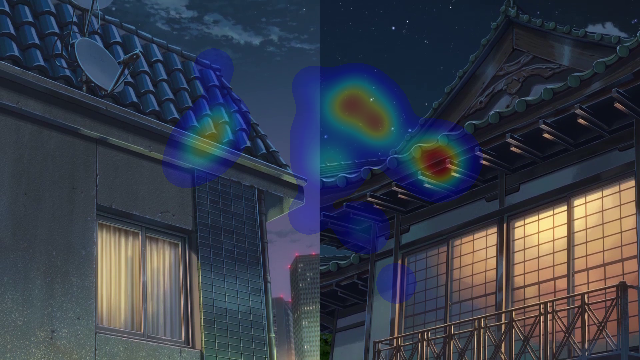}\\
                \includegraphics[width=\linewidth]{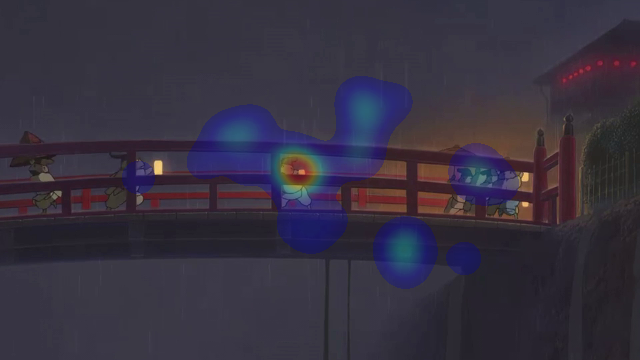}\\
                \includegraphics[width=\linewidth]{figure/experiment/results/3/GT_71_196.png}\\
                \includegraphics[width=\linewidth]{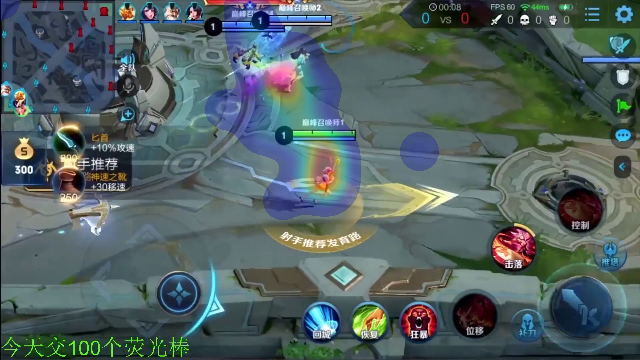}\\
                \includegraphics[width=\linewidth]{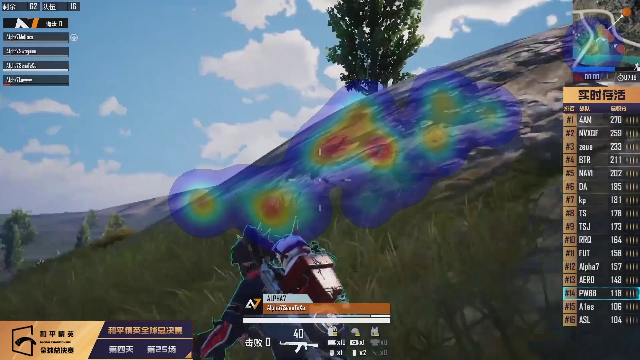}
            \end{minipage}
        }
        \vspace{-3mm}\caption{Visualization results produced by different methods.}
        \label{fig:comparison_result}
\end{teaserfigure}


\maketitle
    
\section{More Experiment Results}
Here we provide more experimental results for evaluating our method. \reftab{ablation} presents the results of different frequencies in cartoon and game domains. {From \reftab{ablation} we can see that high-frequency filters play the most important role, and all three filters are useful. Besides, we provide the computational statistics of different methods in \reftab{complexity}. Our NPSNet outperforms ViNet (previous SOTA) with fewer parameters and computation costs as shown in \reftab{complexity}.}

\begin{table}[t]
\caption{{Ablation study validates that the high-frequency and low-frequency spatial filters contribute mostly.}}
\begin{tabular}{lll}
\hline
\textbf{Model}     & \textbf{Cartoon}  & \textbf{Game}   \\ \hline
{None}      & {1.7811}  & {2.2231} \\ \hline
{Low-frequency}  & {1.8516}  & {2.4122}  \\ \hline
{high-frequency}    & {1.8811}  & {2.5087}  \\ \hline 
{Channel}        & {1.8025} & {2.3972} \\ \hline
{UFM} & {1.9192} & {2.5737} \\ \hline
\end{tabular}
\label{tab:ablation}
\end{table}

\begin{table}[t]
\caption{{Comparison on computational efficiency or model complexity}}
\begin{tabular}{lll}
\hline
\textbf{Model}     & \textbf{Param (M)}  & \textbf{Flops (G)}   \\ \hline
{STSANet \cite{wang2021spatio}}      & {60.0}  & {138.2} \\ \hline
{HD2S \cite{bellitto2021hierarchical}} & {29.8}  & {38.7}  \\ \hline
{TASED-Net \cite{min2019tased}}    & {21.3}  & {78.8}  \\ \hline 
{AViNet \cite{jain2021vinet}}        & {134.2} & {171.2} \\ \hline
{NPSNet (Ours)} & {122.5} & {152.1} \\ \hline
\end{tabular}
\label{tab:complexity}
\end{table}

\section{Visualization Results}
Here we show the visualization results of different methods in Fig.~\ref{fig:comparison_result}. We can see that our method can detect the most salient objects/events in the scenes.
    

{\small
\bibliographystyle{ieee_fullname}
\bibliography{egbib}
}